\pdfoutput=1

\documentclass[11pt]{article}

\usepackage[preprint]{acl}

\usepackage{times}
\usepackage{latexsym}

\usepackage[T1]{fontenc}

\usepackage[utf8]{inputenc}

\usepackage{microtype}

\usepackage{inconsolata}

\usepackage{graphicx}

\usepackage{tikz}
\usepackage{xcolor}
\usepackage{multicol}

\usepackage{stfloats}

\title{Are Large Language Models Chameleons? \\
An Attempt to Simulate Social Surveys}

\author{%
  Mingmeng Geng$^{1}$ \quad Sihong He$^{2}$ \quad Roberto Trotta$^{1,3}$\\
  $^1$International School for Advanced Studies (SISSA) \quad $^2$UT Arlington \\ $^3$Imperial College London \\
  \texttt{mgeng@sissa.it}
}

\usepackage{amsmath}
\usepackage{amssymb}
\usepackage{mathtools}
\usepackage{amsthm}

\usepackage{subcaption}
\usepackage{graphicx}

\usepackage{multirow}
\usepackage{array}
\usepackage{bbm}

\usepackage{booktabs}
\usepackage{tablefootnote}
\begin{document}
\maketitle
\begin{abstract}
Can large language models (LLMs) simulate social surveys? To answer this question, we conducted millions of simulations in which LLMs were asked to answer subjective questions. A comparison of different LLM responses with the European Social Survey (ESS) data suggests that the effect of prompts on bias and variability is fundamental, highlighting major cultural, age, and gender biases. We further discussed statistical methods for measuring the difference between LLM answers and survey data and proposed a novel measure inspired by Jaccard similarity, as LLM-generated responses are likely to have a smaller variance. Our experiments also reveal that it is important to analyze the robustness and variability of prompts before using LLMs to simulate social surveys, as their imitation abilities are approximate at best.
\end{abstract}

\section{Introduction}
Although often referred to as ``stochastic parrots'' \citep{bender2021dangers}, large language models (LLMs) can do more than simply ``talk'' \citep{achiam2023gpt,touvron2023llama,jiang2023mistral}. Their adaptability and versatility remind us of another animal: the chameleon. 

Starting from the very first roles of this LLMs boom - chatbots - can we simulate social surveys through question-and-answer interactions with LLMs? In survey research, several demographic variables are routinely used for sampling, such as gender, race, and age \citep{goldberg1998demographic,herek2002gender,kyriakidis2015public}. Even before the LLMs frenzy began, researchers were already talking about the human values presented in language models \citep{hendrycks2020aligning,santurkar2023whose}.

Researchers also found that LLMs can infer people's gender, age, and location from redacted and anonymized input text \citep{staab2023beyond}. Conversely, prompting LLMs with appropriate demographic variables as we have done here produces answers that appear to be generally matched with real people's views, although still suffering from biases and stereotypes.

Bias in LLMs is almost impossible to avoid, even with subsequent adjustments and alignments \citep{navigli2023biases}. Some researchers have also tried to include demographic variables in the prompt for LLMs. Gender and race are two important perspectives to consider for bias in LLMs \citep{sun2023aligning,kotek2023gender,dong2024disclosure}. The sensitivity of LLMs to age was also analyzed, for example, based on six age groups \citep{liu2024generation}.

There are almost certainly differences in opinions between people and LLMs, but the key is where and how big the gaps are. Furthermore, biases and stereotypes present in LLMs are likely to slowly influence their users \citep{rottger2024political}, so it is essential to measure them. 

Prompts affect the output of LLMs, and the demographic variables in the surveys above can also be viewed as part of the prompt engineering. However, the influence and robustness of the prompts are rarely discussed in papers dealing with opinions expressed by LLMs \citep{rottger2024political}. The ability of LLMs to simulate surveys needs to be explored in more depth.

There are usually no standard answers to public opinion in survey questionnaires. Comparing the responses of LLMs to subjective questions with those provided by human beings can be achieved in multiple ways \cite{durmus2023towards}. Importantly, human opinions are not distributed regularly among people.

LLMs such as ChatGPT may also be more familiar with American culture, compared to other countries \cite{cao2023assessing}. While most research has focused on the United States, our concern is with the European countries, which are more heterogeneous in terms of culture and languages. Therefore, the generalization ability of LLMs can be better compared using European datasets.

In the face of these challenges and gaps, our objectives and contributions are mainly as follows:
\begin{itemize}
    \item Comparison of LLMs responses with real survey data, especially for Europeans;
    \item Comparing the effects of different models and analyzing the impact of prompts using appropriate comparison criteria;
    \item Suggestions and caveats for simulating social surveys using LLMs.
\end{itemize}

LLMs have demonstrated the ability and potential to transform the field of computational social science \citep{ziems2024can}, including survey research \citep{jansen2023employing}. However, some more fundamental issues need to be validated and addressed, which is the main target of this paper.

\section{Related work}

\paragraph{LLM simulation} People have also used LLMs to create ``silicon samples'' for simulations, from GPT-3 \cite{argyle2023out,santurkar2023whose} to other newer models \cite{takemoto2024moral,boelaert2024machine}. What could be called the ``personality traits'' of LLMs have also been widely explored in several papers \citep{safdari2023personality,jiang2023personallm,jiang2024evaluating,mei2024turing}. While most of the papers are based on American data \cite{bisbee2023synthetic,sanders2023demonstrations}, very few specifically relate to Europeans. Most researchers focus on the mean of the LLMs simulation and few studies pay attention to the variance \cite{boelaert2024machine}.

\paragraph{LLM bias} Bias in natural language processing (NLP) arises for several reasons, such as data, models, and research design \citep{hovy2021five}. The risks and shortcomings of LLMs were addressed even before they became popular \citep{bender2021dangers}. While LLMs solve many of the difficulties that traditional NLP methods have failed to overcome, the challenge of bias persists \citep{navigli2023biases}. For example, results based on GPT-3 show that the opinions of certain demographic groups in the United States are not well represented \citep{santurkar2023whose}. The political bias of ChatGPT in other countries has also been quantified \citep{motoki2024more}. Bias also looks like a problem shared by other LLMs \citep{boelaert2024machine}, and not specific to ChatGPT. 

\paragraph{LLM evaluation} The pervasiveness and influence of LLMs is increasing rapidly, and it is becoming more important to better evaluate and compare different LLMs \citep{chang2024survey}. Platforms that build on users' comparative ratings have also emerged, for example, \textit{Chatbot Arena} \citep{chiang2024chatbot}. Traditional objective evaluation criteria for solving various tasks, such as MMLU (Massive Multitask Language Understanding) \citep{hendrycks2020measuring}, do not meet current needs. Therefore, novel evaluation methods have been proposed, such as uncertainty quantification \citep{ye2024benchmarking}.

\paragraph{LLM alignment} How to align LLMs with human behavior is an exciting and challenging task \citep{ouyang2022training}. Despite the many limitations \citep{wolf2023fundamental,kirk2023personalisation}, we can see the attempts and results of many researchers on the alignment of LLMs \citep{shen2023large,liu2023trustworthy}. LLMs are used in a variety of applications, so corresponding datasets and benchmarks are also needed for different usage scenarios \citep{kopf2024openassistant,koo2023benchmarking}.

\paragraph{Prompt engineering} Prompts affect the output of LLMs \cite{white2023prompt}, such as the language used \cite{agarwal2024ethical,alkhamissi2024investigating}, the order of inputs in the in-context learning setting \cite{lu2021fantastically}, the order of the options in the question \cite{hartmann2023political,dominguez2023questioning}, and of course the demographic variables included \cite{beck2024sensitivity}.

\section{Data}
The data and results obtained from surveys also deviate to a greater or lesser extent from the ``ground truth'', while the representativeness of the sample is often debated \citep{russell2022any,grafstrom2014select}. Therefore, we want to analyze the bias in the responses of LLMs, using real survey data as the ``ground truth'', which means that the survey results are assumed to be representative. Therefore, we use survey data rather than crowd-sourced data as a benchmark.

To better compare with real people's opinions, we made use of data from the European Social Survey (ESS)\footnote{\url{https://www.europeansocialsurvey.org/}}, a biennial survey of attitudes and behavior. We chose the round 10 of ESS \citep{European_Social_Survey_ERIC_ESS_ERIC2022-nn} (details in the appendix)  with 59,686 participants from 31 European countries, released from June 2022 to May 2023. Germany has the most participants, totaling 8,725, while Cyprus has the least, with only 875 participants.

\section{Methods}
\subsection{Simulations settings}
The performance of LLMs needs to be evaluated from simulations, as it is not possible to predict the response of LLMs through analysis of the model structure and its parameters. 

\paragraph{Models} We investigated ChatGPT (including \textit{GPT-3.5} and \textit{GPT-4o}), as well as open-source models via APIs or locally, including LLaMA-2 \citep{touvron2023llama}, LLaMA-3, Mistral \citep{jiang2023mistral} and DeepSeek-V2 \citep{bi2024deepseek}. As with the real survey, we used the \textit{Chat} model rather than the completion of sentences. 

\paragraph{Prompts} The prompts are set as in Figure \ref{prompt}, where the underlined information is obtained from real survey data rather than randomized settings. For the sake of comparison, we will only consider questions with these five options below in the survey. 
\begin{figure}[h]
    \centering
    \begin{tikzpicture}
        \fill[blue!20] (-3.8, 2) rectangle (3.8, 5);  
        \node at (0, 3.5) {\parbox{6.8cm}{
            \centering
            \textbf{Prompt} You are a \underline{man} born in \underline{1990} living in \underline{Germany}. \textit{Occupation category: \underline{Waiters}.} To what extent do you agree with the following statement: `[\underline{statement}]'? \\
            Please respond with only a number
        }};

        \fill[green!20] (-3.8, -1.2) rectangle (0, 2); 
        \node[anchor=north west] at (-3.8, 1.9) {\parbox{3.3cm}{
            (1) Agree strongly\\
            (2) Agree\\
            (3) Neither agree nor disagree\\
            (4) Disagree\\
            (5) Disagree strongly.}};

        \fill[yellow!20] (0, -1.2) rectangle (3.8, 2); 
        \node[anchor=north west] at (0.2, 1.9) {\parbox{3.3cm}{
            (1) Disagree strongly\\
            (2) Disagree\\
            (3) Neither agree nor disagree\\
            (4) Agree\\
            (5) Agree strongly.
                
        }};

        \draw[thick] (-3.8, -1.2) rectangle (3.8, 5);
    \end{tikzpicture}
    \caption{LLM simulation with different prompts.}
    \label{prompt}
\end{figure}

\paragraph{Responses} We ask one question at a time, so each answer is zero-shot. Responses from LLMs were also diverse and sometimes LLMs avoided answering, with examples given in Section \ref{responses} of the Appendix. 

\paragraph{Simplifications} For questions offering the five options above, we focus on data where participants agree or disagree with a statement, while for simplicity the missing data options (``Refusal'', ``Don't know'', ``No answer'') were disregarded.

\paragraph{Parameters} In addition to the different prompts, the influence of parameters cannot be ignored, such as \textit{temperature} and \textit{top\_p}. Different parameters were tested to compare the effects of the models.

\subsection{Measurements}
The literature tells us that LLMs are usually biased, but there is no agreement on how large these biases are. As previously discussed, there are many ways to measure this bias, with varying conclusions.

Unlike many question-and-answer type tasks with LLMs, social surveys do not have standardized correct answers, but do have references to the distribution of real data. In orde to compare responses from LLMs to the ones from humans, we want to evaluate not only averages but also their distributions, as it's not binary classification. 

Inspired by Jaccard similarity \citep{chierichetti2010finding}, we define the J-index $J_q(G)$ for statement $q$ and people group $G$ that we are interested in (e.g., a country) as:
\begin{equation}
        J_q(G) = \frac{\sum_{g \in G}I\left(R^h_q(g), R^m_q(g))\right)}{\sum_{g \in G}U\left(R^h_q(g), R^m_q(g))\right)}
\end{equation}
where $I\left(\cdot,\cdot\right)$ and $U\left(\cdot,\cdot\right)$ are the intersection and union of two sets of responses, $g$ is the subgroup of $G$, $R^h_q(g)$ and $R^m_q(g)$ represent the responses from survey data and LLM simulations, respectively. A J-index of 1 thus corresponds to perfect congruence between between survey and LLM simulation.

Taking advantage of the data from the survey, $I\left(R^h_q(g), R^m_q(g)\right)$ and $U\left(R^h_q(g), R^m_q(g)\right)$ are defined as follows,
\begin{align*}
\begin{split}
    &I\left(R^h_q(g), R^m_q(g)\right) \\ =&\sum_{v\in V_q} \left( \min \left( \sum_{i \in g} w_i\mathbbm{1}_{r^h_q(i)=v},\sum_{i \in g} w_i\mathbbm{1}_{r^m_q(i)=v} \right) \right)
\end{split}\\
\begin{split}
     &U\left(R^h_q(g), R^m_q(g)\right) \\ =&
    \sum_{v\in V_q} \left( \max \left( \sum_{i \in g} w_i\mathbbm{1}_{r^h_q(i)=v},\sum_{i \in g} w_i\mathbbm{1}_{r^m_q(i)=v} \right) \right)   
\end{split}
\end{align*}
where $v$ corresponds to the value of the response, $V_q$ is the set of all possible values for question $q$, $r^h_q(i)$ represents the answer in survey data, $r^m_q(i)$ means the response of LLM given the same information as participant $i$, and $w_i$ is the analysis weight (combination of post-stratification weight and population weight) provided in the ESS dataset, which could improve the representativeness of survey respondents concerning the target population.

The measure we defined above allows for a better comparison of how similar the distributions are, bringing in some information different from the mean.

The ordinal numbers of the responses from $1$ (``Agree strongly'') to $5$ (``Disagree strongly'') were used as a scale to calculate the mean of their disagreement level.

\section{Results}
In our analysis, we used the 4 countries with the most participants in the ESS (Germany, Greece, Bulgaria, and Italy, with 16132 samples in total). All of our results take post-stratification weighting into account.

\subsection{ChatGPT simulations}

As described in Figure \ref{prompt}, we tested the chameleon ability of LLM given birth year, gender, and residence information. The specific birth years rather than age groupings were adopted to better show the sensitivity of LLM to age information. The first prompt \textbf{P1} is shown below.

\paragraph{P1} \textit{You are a [man or woman] born in [year] living in [country]. To what extent do you agree with the following statement: `[statement]'? Please respond with only a number (1) Agree strongly (2) Agree (3) Neither agree nor disagree (4) Disagree (5) Disagree strongly.}
\begin{figure}[t]
    \centering
    \begin{subfigure}[b]{0.45\textwidth}
        \includegraphics[width=\textwidth]{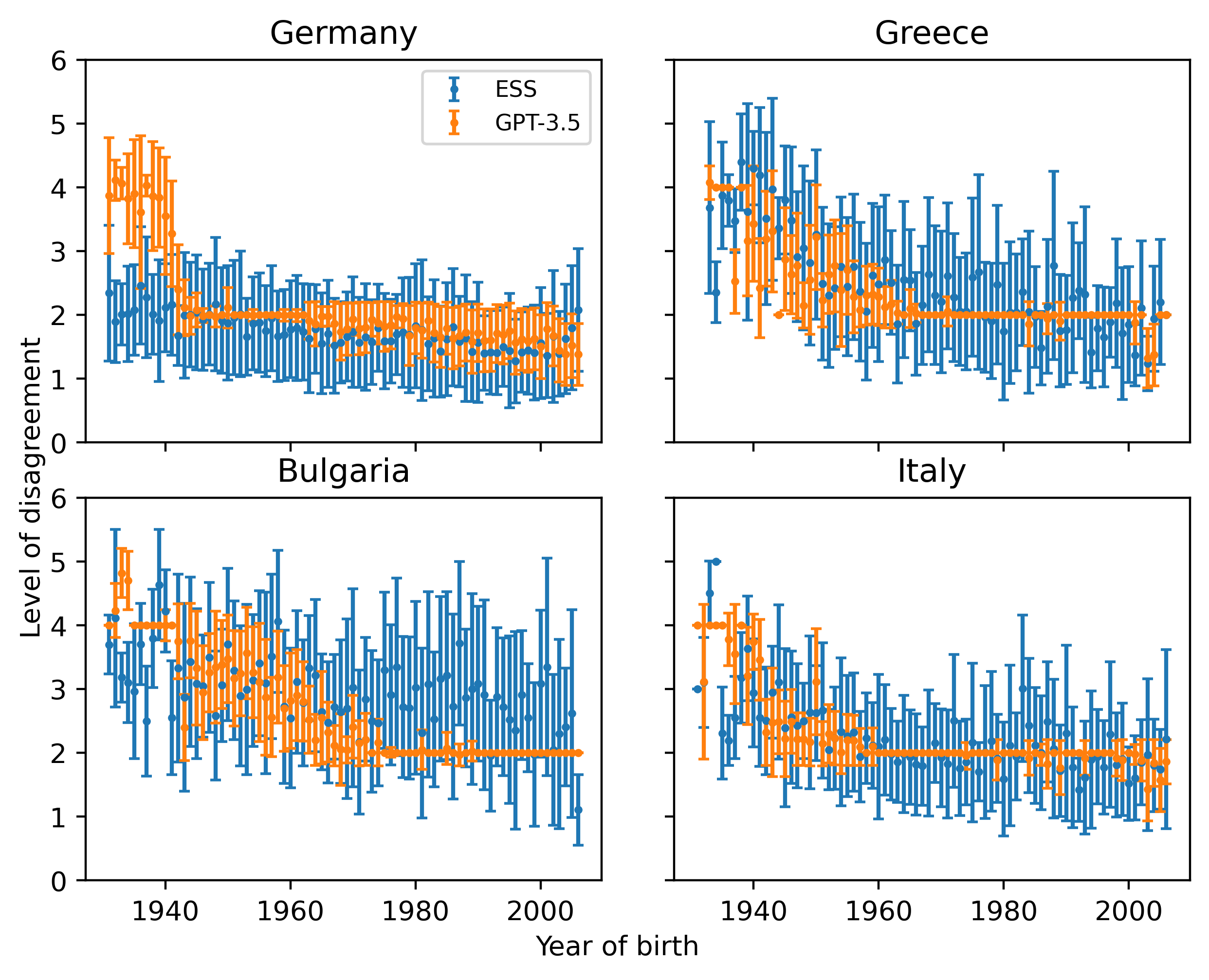}
        \caption{Men}
    \end{subfigure}
    \hfill
    \begin{subfigure}[b]{0.45\textwidth}
        \includegraphics[width=\textwidth]{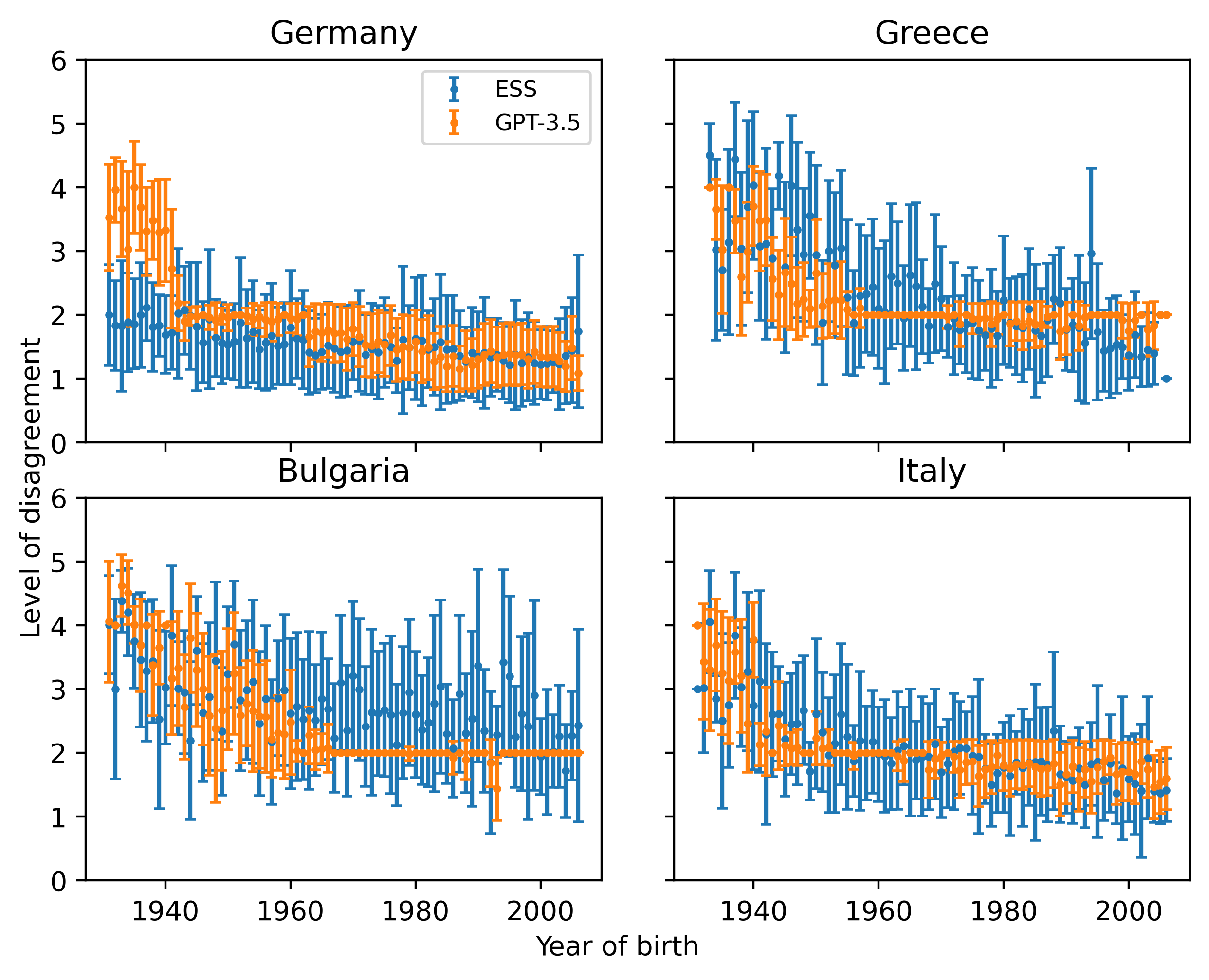}
        \caption{Women}
    \end{subfigure}
    \caption{\textit{Q1}: ``Gays and lesbians free to live life as they wish''? Prompt: \textbf{P1}. The points represent the mean and the error bars represent the standard deviation (and the same for the next figures). Model: \textit{GPT-3.5}.}
    \label{freehms}
\end{figure}

The results for the first statement (\textit{Q1}) with \textit{GPT-3.5} \footnote{gpt-3.5-turbo-0125, $\textit{temperature}=1$, $\textit{top\_p}=0.9$} are shown in Figure \ref{freehms}, where simulated responses and real survey data are consistent. ChatGPT does not always respond with only one number as we requested, sometimes including a description after the number (e.g., ``(1) Agree strongly''), but it doesn't refuse to answer.

The average of the simulation results is quite close to the mean of ESS data, except for Germans born before 1940. However, the variance of the simulated data is too small compared to the real ones.

\begin{figure}[t]
    \centering
    \begin{subfigure}[b]{0.45\textwidth}
        \includegraphics[width=\textwidth]{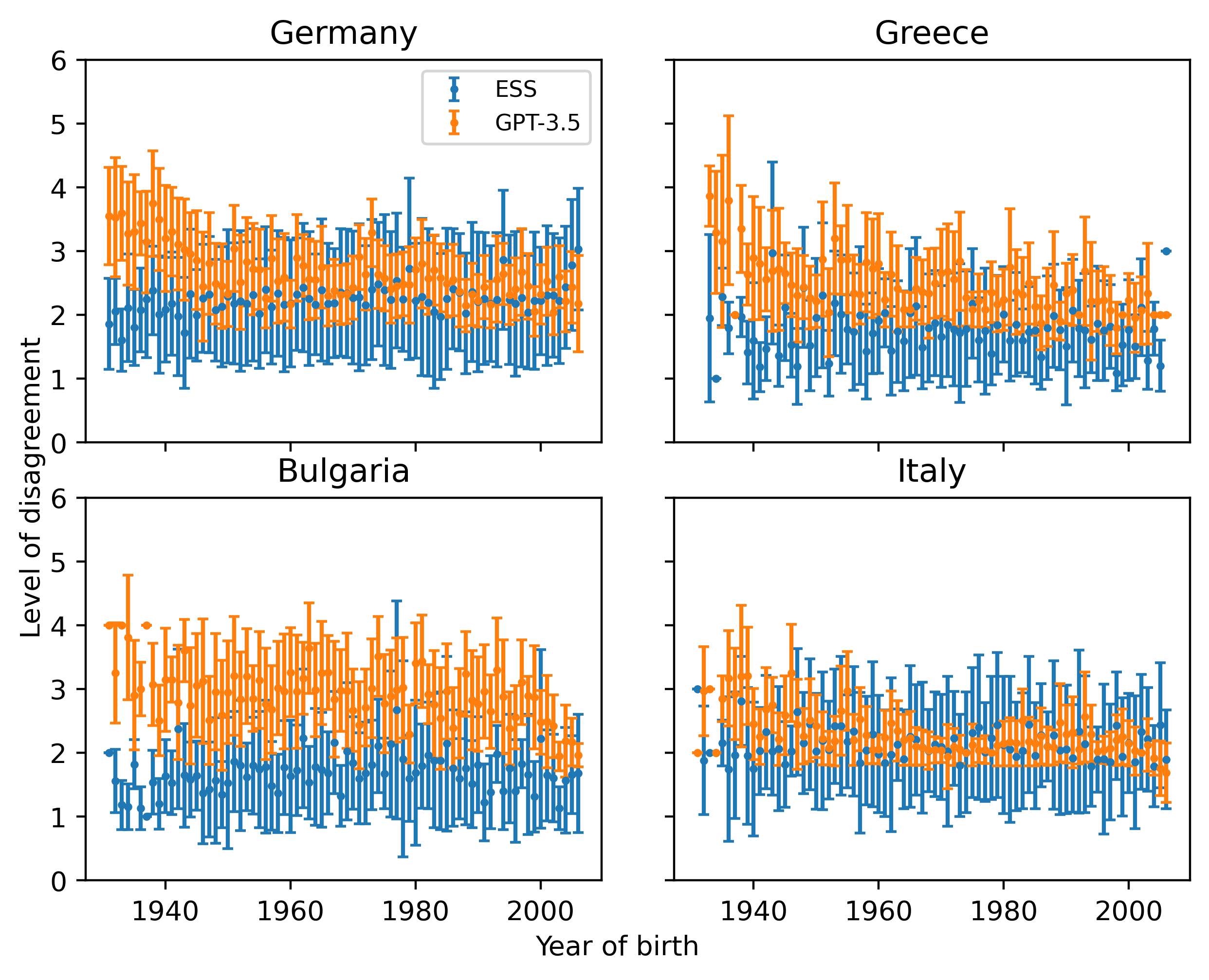}
        \caption{Men}
    \end{subfigure}
    \hfill
    \begin{subfigure}[b]{0.45\textwidth}
        \includegraphics[width=\textwidth]{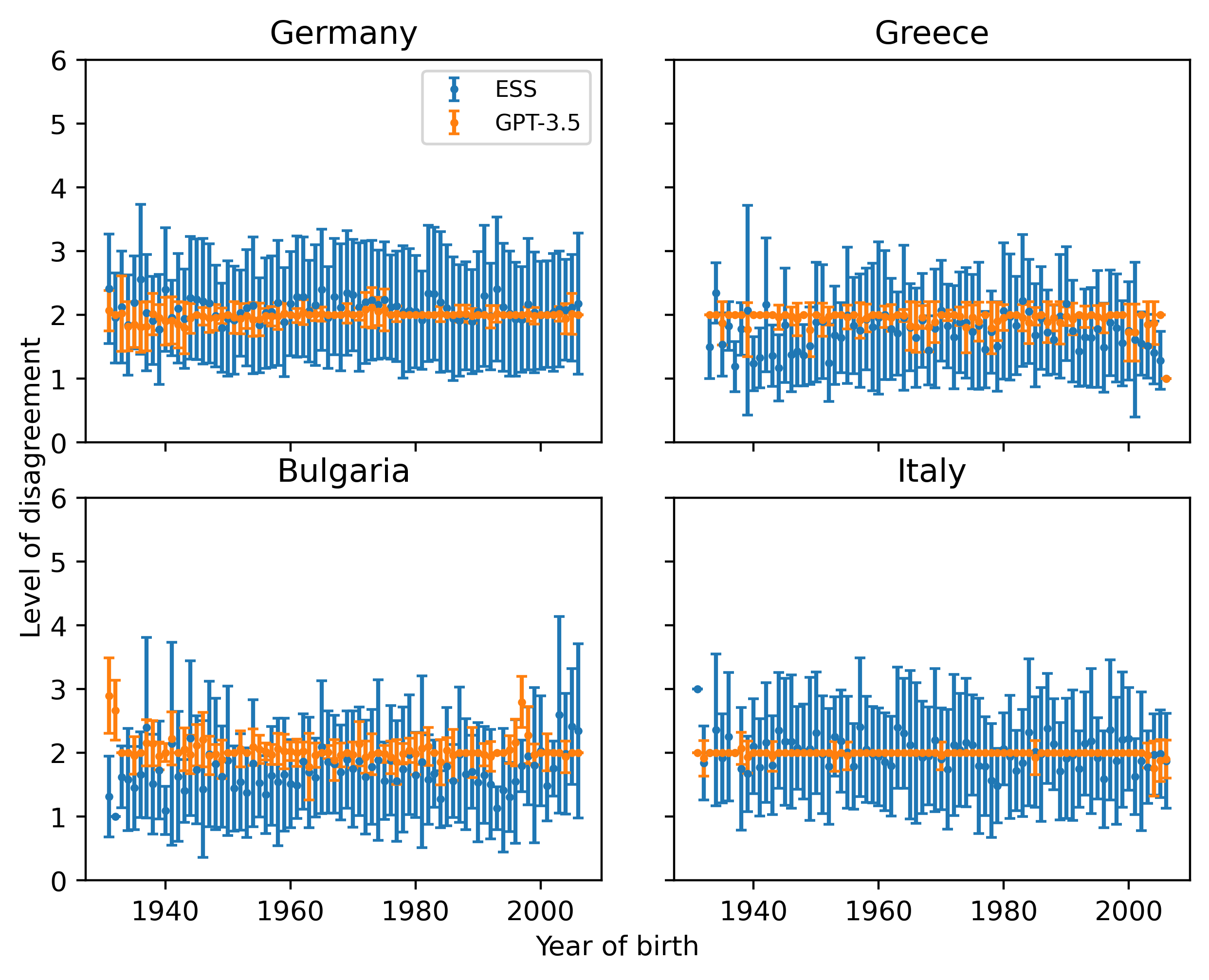}
        \caption{Women}
    \end{subfigure}
    \caption{\textit{Q2}: ``Government should reduce differences in income levels''? Prompt: \textbf{P1}. Model: \textit{GPT-3.5}.}
    \label{gincdif}
\end{figure}

The impact of the same variable can vary for different topics, e.g, the difference resulting from gender is evident in LLMs' response to the second question, as seen in Figure \ref{gincdif}. From here we can see  better the importance of considering the distribution rather than the mean in some cases.

Although we always used different random seeds for different persons and the same random seed for the same person across different questions, the problem of too little variance in the simulations is common for various questions, as has been pointed out by others \citep{boelaert2024machine}. 

In particular, the example in Figure \ref{gincdif} shows that although the mean of ESS data and \textit{GPT-3.5} simulations are close for women, the variance of simulations is very small --almost non-existent-- in this case. This would seem to imply that women are associated with more stereotypical answers, another facet of gender bias. The results of such simulations can hardly be described as reproducing the social survey data.

\subsection{Prompts}

\begin{figure}[t]
    \centering
    \begin{subfigure}[b]{0.45\textwidth}
        \includegraphics[width=\textwidth]{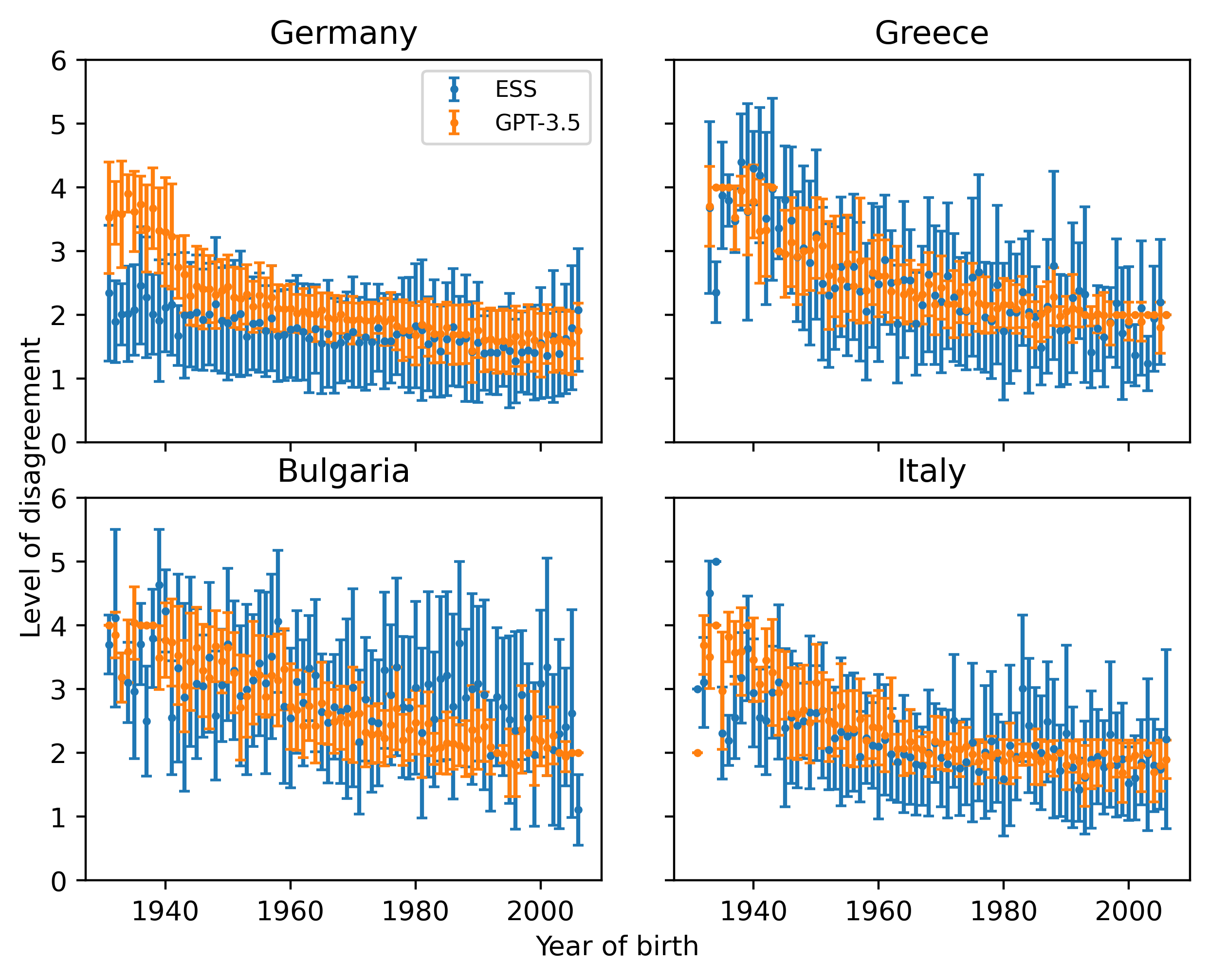}
        \caption{Men}
    \end{subfigure}
    \hfill
    \begin{subfigure}[b]{0.45\textwidth}
        \includegraphics[width=\textwidth]{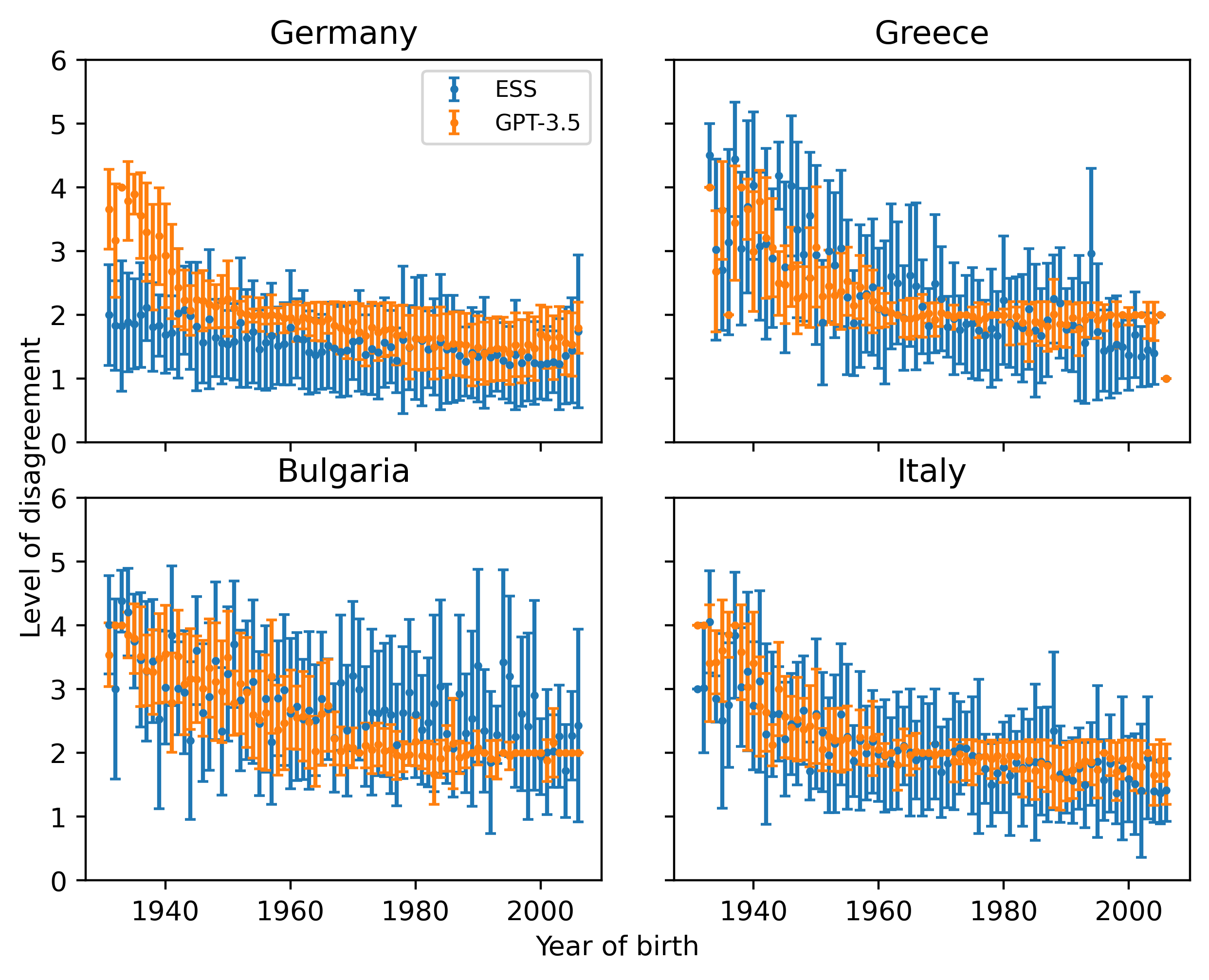}
        \caption{Women}
    \end{subfigure}
    \caption{\textit{Q1}: ``Gays and lesbians free to live life as they wish''? Prompt: \textbf{P2}. Model: \textit{GPT-3.5}.}
    \label{freehms_isco}
\end{figure}

Intuitively, simulations containing the same demographic variables are considered the ``same people'' by LLM, so the variance of LLM responses is smaller. But the corresponding actual survey respondents are quite different. Therefore, we tried prompts with a more informative description, such as the respondent's occupation:

\paragraph{P2} \textit{You are a [man or woman] born in [year] living in [country]. \textbf{Occupation category: [ISCO\footnote{International Standard Classification of Occupations (ISCO)}]}.  To what extent do you agree with the following statement: `[statement]'? Please respond with only a number (1) Agree strongly (2) Agree (3) Neither agree nor disagree (4) Disagree (5) Disagree strongly.}

The results generated by prompt \textbf{P2} are shown in Figure \ref{freehms_isco}, where the variability of simulations becomes closer to the variance of survey data. This is not because our addition to the prompt has increased the degree of randomness, but rather because occupational differences were previously ignored, suggesting that a ``regression to the mean'' phenomenon was at play with prompt \textbf{P1}.

The impact of prompts goes beyond how much information is provided -- the order also matters, as has been shown previously \citep{pezeshkpour2023large}. For example, we tried reversing the order of answer options: 
\paragraph{P3} \textit{You are a [man or woman] born in [year] living in [country]. To what extent do you agree with the following statement: `[statement]'? Please respond with only a number \textbf{(1) Disagree strongly (2) Disagree (3) Neither agree nor disagree (4) Agree (5) Agree strongly.}}

Comparing the corresponding simulation results for \textbf{P1} and \textbf{P3}, the prompts can sometimes seriously affect the output, as shown in Figure \ref{freehms_inv} in the appendix. The simulation results contain more ``\textit{Agree strongly}'' and less ``\textit{Agree}'' for \textit{Q1} compared to Figure \ref{freehms}, although the overall trend is still consistent.

As the order in which options are present affects humans' choice \citep{galesic2008eye}, we can't state which LLM mimics humans more based on how robust it is to the order of options. The same appears to be true for LLMs, to an even larger extent at times. Comparisons should be made with the fluctuations of human choices, although this is not addressed in this paper.

\begin{figure}[t]
    \centering
    \begin{subfigure}[b]{0.45\textwidth}
        \includegraphics[width=\textwidth]{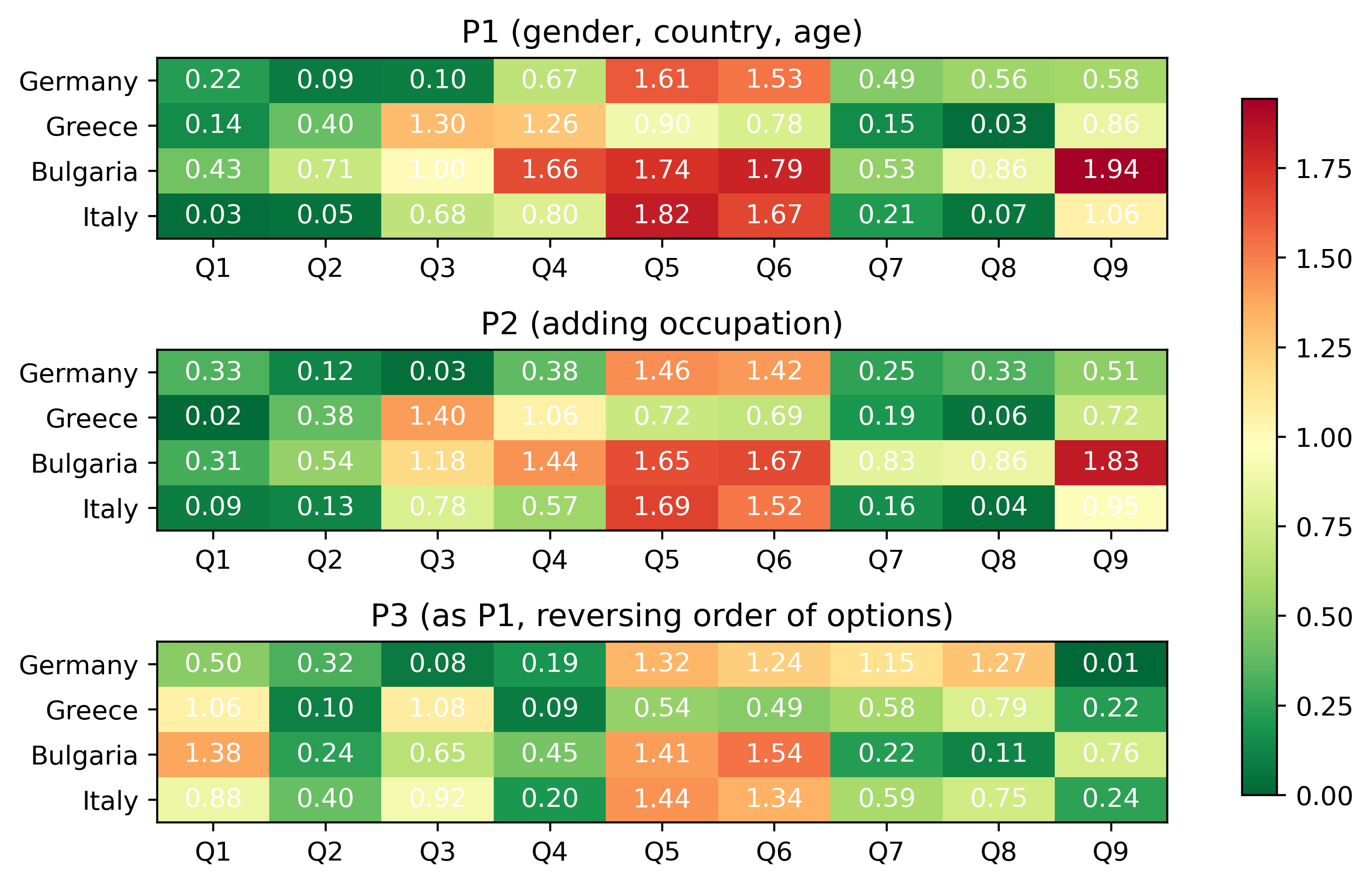}
        \caption{Absolute value of the mean difference (error).}
    \end{subfigure}
    \hfill
    \begin{subfigure}[b]{0.45\textwidth}
        \includegraphics[width=\textwidth]{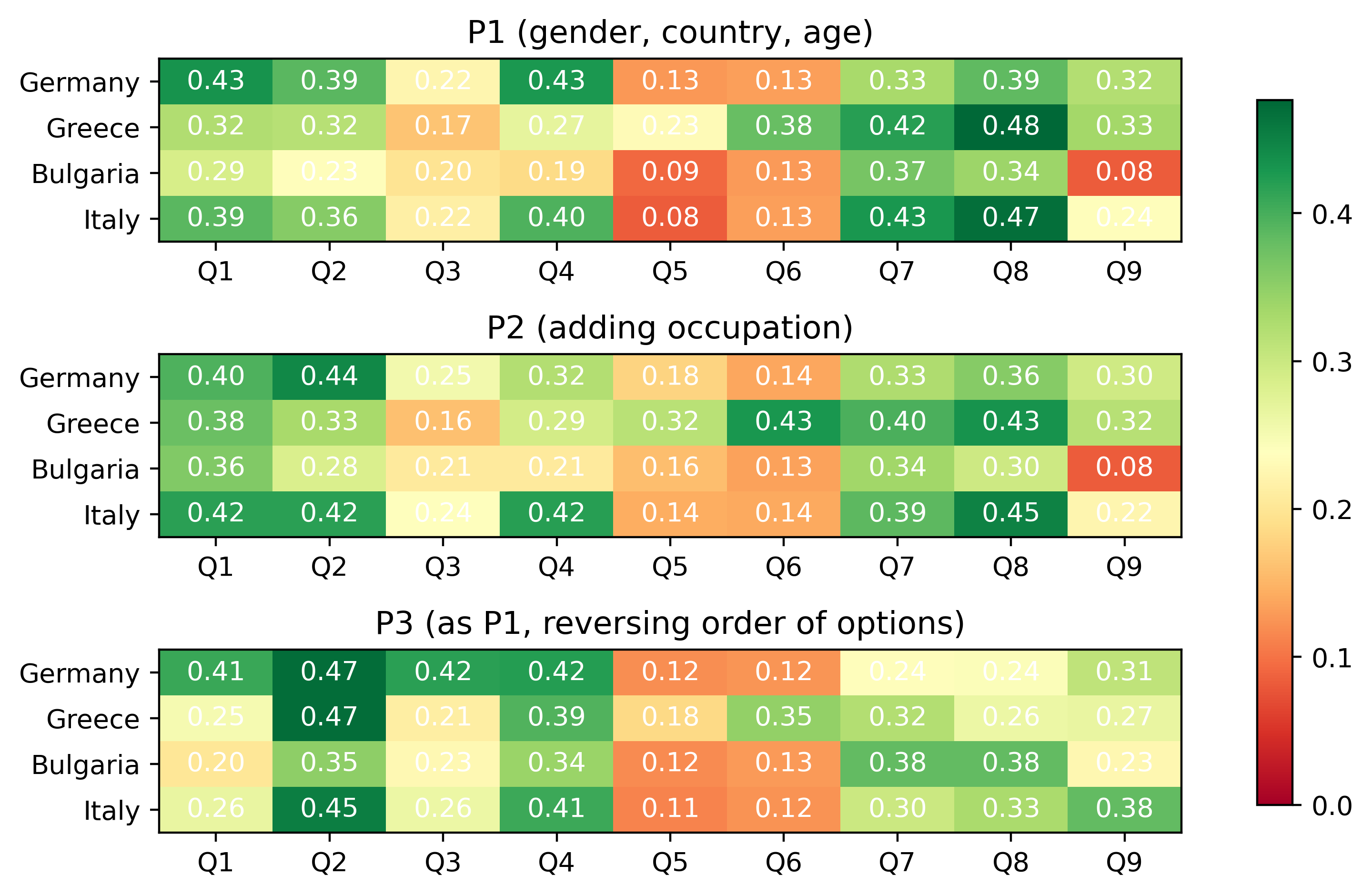}
        \caption{J-index (congruence). }
    \end{subfigure}
    \caption{Comparisons between survey data and simulation results based on \textit{GPT-3.5}.}
    \label{res_gpt}
\end{figure}

For 9 such questions in the ESS (listed in the appendix), we calculated the mean bias and J-index for the simulations with \textit{GPT-3.5} compared to ESS data, and the results are shown in Figure \ref{res_gpt}. For ease of plotting, we take the absolute value of the mean difference.

\paragraph{Mean difference} As we have seen above, prompt \textbf{P3} leads to markedly different simulation results for some questions, like \textit{Q1} and \textit{Q4}, but not for others. Moreover, the simulation results for \textit{Q4} using prompt \textbf{P3} produced better results than those using prompts \textbf{P1}, while the opposite is true for \textit{Q1}. The order of the options affects the results of the LLMs simulations, which vary in sensitivity from question to question, and it does not appear to be possible to predict which order will result in results closer to the real survey data.

\paragraph{J-index} The J-index brings further insight. When more personal information is included in the prompts (i.e., \textbf{P2}), the bias decreases for 22 out of 36 and the J-index increases for 25 out of 36 cases. The results also illustrate that adding personal information will have a greater chance of making the LLMs closer to the real survey data, in terms of mean and distribution.

\subsection{Imbalance and bias}

\begin{figure}[t]
    \centering
    \begin{subfigure}[b]{0.45\textwidth}
        \includegraphics[width=\textwidth]{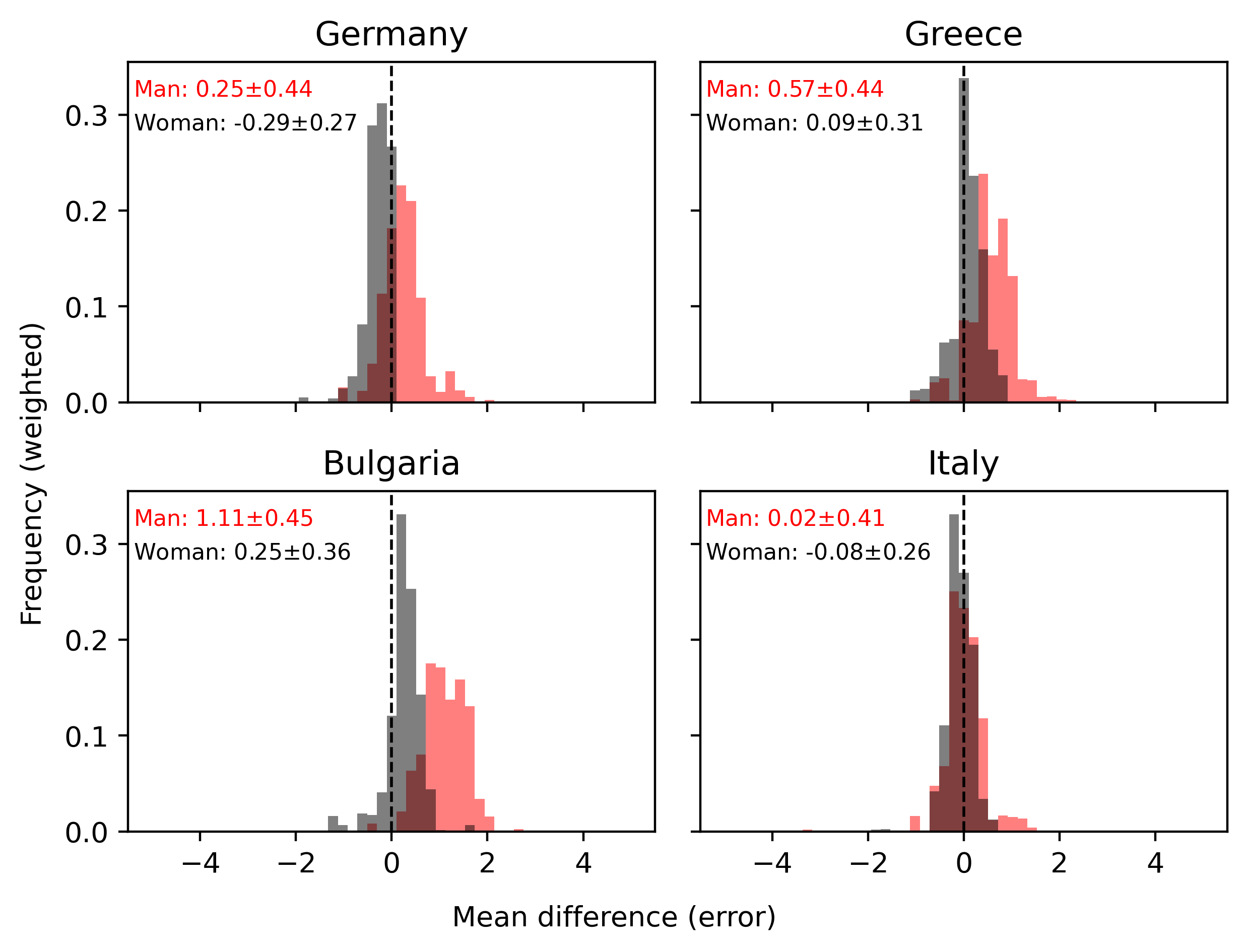}
        \caption{Model: \textit{GPT-3.5}}
    \end{subfigure}
    \hfill
    \begin{subfigure}[b]{0.45\textwidth}
        \includegraphics[width=\textwidth]{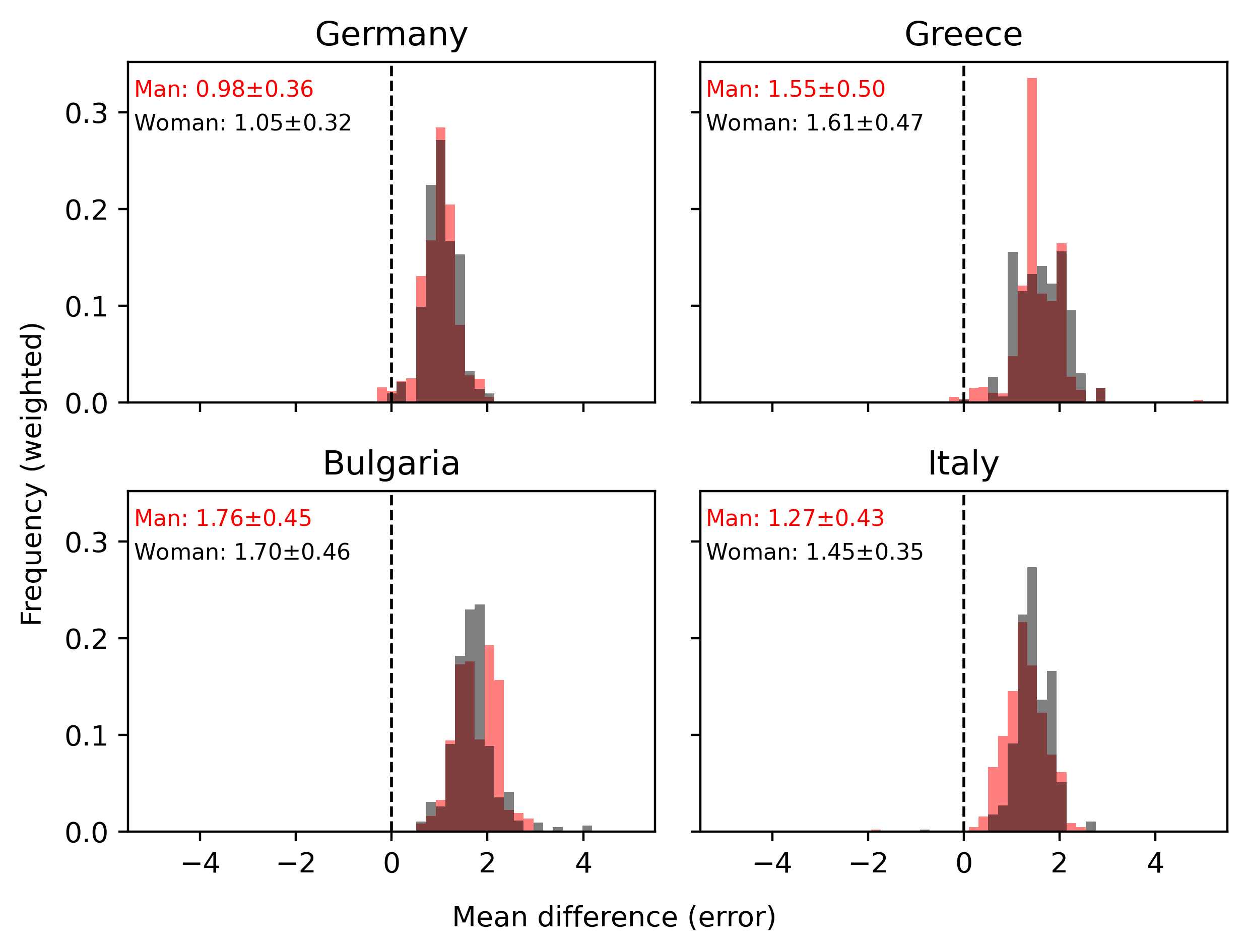}
        \caption{Model: \textit{LLaMA-2-7B}}
    \end{subfigure}
    \caption{\textit{Q2}: ``Government should reduce differences in income levels''? Prompt: \textbf{P1}.}
    \label{gincdif_compare}
\end{figure}

Regardless of the goals and process of LLM alignment, our results reflect a clear geographic imbalance. Simulations of Bulgarians work worse compared to people from other countries, for reasons that remain to be explored. Our guess is that the training and fine-tuning process of LLMs uses less data about Bulgarians. 

Figure \ref{gincdif_compare} compares the mean difference of each gender for \textit{Q2} between \textit{GPT-3.5} and \textit{LLaMA-2-7B}, where 
people with the same country, gender, and year of birth are considered to be in the same group. While \textit{LLaMA-2-7B} shows a larger bias, it appears to be almost gender-independent.  \textit{GPT-3.5} presents an overall smaller bias, but a more marked influence of gender in both mean and standard deviation.

\subsection{Comparisons between different LLMs}

\begin{table*}[b]
\centering
  \caption{Results from different LLMs (mean difference between LLMs answers and survey). Values in bold are the closest simulations to the mean for each row, while those in red italics are the worst. LLMs: \textit{GPT-3.5}, \textit{GPT-4o}, \textit{LLaMA-2-7B} (L-7B), \textit{LLaMA-2-13B} (L-13B), \textit{LLaMA-3-8B} (L-8B), \textit{Mistral-7B-Instruct-v0.2} (Mistral), \textit{DeepSeek-V2} (DS).}
  \label{res_llms_md}
\begin{tabular}{c c c c c c c c c}
\toprule
\textbf{country} & \textbf{prompt} & \textbf{GPT-3.5} & \textbf{GPT-4o} & \textbf{L-7B} & \textbf{L-13B} & \textbf{L-8B} & \textbf{Mistral} & \textbf{DS} \\ \midrule
\multirow{2}{*}{\centering Germany} & \textbf{P1} & 0.22 & \textcolor{red}{\textit{1.37}} & 0.51 & 0.64 & 1.19 & 0.47 & \textbf{-0.12} \\ \cmidrule{2-9}
                       &\textbf{P2} & 0.33 & 0.33 &  0.40 & 0.57 & \textcolor{red}{\textit{1.48}} & \textbf{-0.13} & -0.29 \\ \midrule
\multirow{2}{*}{Greece} & \textbf{P1} & -0.14 & \textcolor{red}{\textit{1.02}} & \textbf{0.04} & 0.21 & 0.75 & 0.86 & -0.19 \\ \cmidrule{2-9}
                       & \textbf{P2} & -0.02 & \textbf{-0.01} & -0.13 & 0.41 & \textcolor{red}{\textit{0.92}} & 0.78 & 0.03 \\ \midrule
\multirow{2}{*}{Bulgaria} & \textbf{P1} & -0.43 & \textcolor{red}{\textit{1.11}} & -0.25 & 0.44 & 0.61 & 1.01 & \textbf{-0.13} \\ \cmidrule{2-9}
                       & \textbf{P2} & -0.31 & \textbf{0.10} & -0.82 & 0.18 & 0.61 & \textcolor{red}{\textit{1.19}} & -0.16 \\ \midrule
\multirow{2}{*}{Italy} & \textbf{P1} & \textbf{-0.03} & 0.76 & 0.24 & 0.30 & \textcolor{red}{\textit{0.94}} & 0.77 & -0.13 \\ \cmidrule{2-9}
                       & \textbf{P2} & 0.09 & -0.32 & \textbf{0.01} & 0.51 & \textcolor{red}{\textit{1.10}} & 0.37 & -0.9 \\ \midrule
\multirow{2}{*}{Avg. (abs.)}& \textbf{P1} & 0.21 & \textcolor{red}{\textit{1.06}} & 0.26 & 0.40 & 0.87 & 0.77 & \textbf{0.14} \\ \cmidrule{2-9}
& \textbf{P2} & 0.19 & 0.19 & 0.34 & 0.42 & \textcolor{red}{\textit{1.03}} & 0.62 & \textbf{0.15} \\
\bottomrule
\end{tabular}
\end{table*}

\begin{table*}[t]
\centering
  \caption{J-index results for different LLMs. Values in bold are the largest $J$-index for each row, while those in red italics are the worst.}
  \label{res_llms_jac}
\begin{tabular}{c c c c c c c c c}
\toprule
\textbf{country} & \textbf{prompt} & \textbf{GPT-3.5} & \textbf{GPT-4o} & \textbf{L-7B} & \textbf{L-13B} & \textbf{L-8B} & \textbf{Mistral} & \textbf{DS} \\ \midrule
\multirow{2}{*}{\centering Germany} & \textbf{P1} & 0.43 & \textcolor{red}{\textit{0.13}} & 0.35 & 0.17 & 0.25 & \textbf{0.51} & 0.46 \\ \cmidrule{2-9}
                       &\textbf{P2} & 0.40 & 0.40 & 0.40 & 0.28 & \textcolor{red}{\textit{0.18}} & 0.47 & \textbf{0.50} \\ \midrule
\multirow{2}{*}{Greece} & \textbf{P1} & \textbf{0.32} & 0.24 & 0.27 & 0.23 & 0.26 & \textcolor{red}{\textit{0.20}} & 0.27 \\ \cmidrule{2-9}
                       & \textbf{P2} & \textbf{0.38} & \textbf{0.38} & 0.26 & \textcolor{red}{\textit{0.15}} & 0.25 & 0.18 & 0.27 \\ \midrule
\multirow{2}{*}{Bulgaria} & \textbf{P1} & \textbf{0.29} & 0.20 & 0.26 & \textcolor{red}{\textit{0.17}} & 0.23 & 0.19 & 0.19 \\ \cmidrule{2-9}
                       & \textbf{P2} & \textbf{0.36} & \textbf{0.36}  & 0.21 & \textcolor{red}{\textit{0.15}} & 0.28 & 0.19 & 0.20 \\ \midrule
\multirow{2}{*}{Italy} & \textbf{P1} & \textbf{0.39} & 0.28  & 0.27 & 0.28 & \textcolor{red}{\textit{0.22}} & 0.37 & 0.33 \\ \cmidrule{2-9}
                       & \textbf{P2} & 0.42 &  0.42 & 0.27 & \textcolor{red}{\textit{0.19}} & 0.20 & \textbf{0.48} & 0.35 \\ \midrule
\multirow{2}{*}{Avg.}& \textbf{P1} & \textbf{0.36} & \textcolor{red}{\textit{0.21}} & \textbf{0.29} & \textcolor{red}{\textit{0.21}} & 0.24 & 0.32 & 0.31\\ \cmidrule{2-9} 
& \textbf{P2} & \textbf{0.39} & \textbf{0.39} & 0.28 & \textcolor{red}{\textit{0.19}} & 0.23 & 0.33 & 0.33 \\
\bottomrule
\end{tabular}
\end{table*}

Different LLMs have different simulation outcomes, and the parameters of the model sometimes play an important role. The information in the prompt and the way it is described can, of course, greatly influence the LLM's response, with no clear indication as to what is the best choice.

For a more quantitative comparison, we calculated the difference of the mean and J-index compared to ESS data for \textit{Q1}, listed in Table \ref{res_llms_md} and Table \ref{res_llms_jac} for 7 different LLMs, using the prompts \textbf{P1} and \textbf{P2}. 

\paragraph{Mean difference} Table \ref{res_llms_md} gives the consistent conclusions as before, showing that \textit{DeepSeek-V2} and \textit{GPT-3.5} generate simulation results with weighted averages closest to the survey data, while \textit{LLaMA-3-8B} is highly biased for both prompts; occupation information is very important for \textit{GPT-4o}, which appears to perform generally worse than its predecessor \textit{GPT-3.5}. While it is somewhat unexpected, \citet{zhou2024larger} also points out the fact that bigger models don't necessarily work better. 

\paragraph{J-index} The conclusions based on Table \ref{res_llms_jac} are very similar to the previous ones, and the results from \textit{Mistral-7B-Instruct-v0.2} are very interesting: the simulation results for Germans and Italians are very close to the real data, while the results of Greeks and Bulgarians pull down the average effect of this model. In a nutshell, we observe that \textit{GPT-3.5} is the best-performing model among the ones tested. 

\paragraph{No response} Unlike ChatGPT, the other LLMs may refuse to answer, and their responses are not always valid (see appendix for examples), a phenomenon that concerns some researchers \cite{zhou2024larger}. We could force them to answer by changing prompts, but this would also affect their choices \citep{rottger2024political}, thus they are considered as missing data. In real surveys, missing data is also a major challenge.

\paragraph{Case study} The results of the simulation on \textit{LLaMA-2-7B} using prompt \textbf{P1} for \textit{Q1} are shown in Figure \ref{freehms_llama_7}, which features a much larger variance compared with \textit{GPT-3.5} shown in Figure \ref{freehms}. The survey data corresponding to questions with invalid responses in the simulations are not used in the calculations.

\subsection{Parameters}
We know that the parameters of the LLMs also have an impact on the results, so we also tested different parameters.

When we set \textit{top\_p} to $0.2$ instead of $0.9$, the J-index becomes worse in all 72 cases with prompts \textbf{P1} and \textbf{P2}. (The specific values are listed in Table \ref{top_p_02} in the appendix.) The bias can both increase or decrease with the smaller \textit{top\_p} value, which is in line with our expectations.

For the parameter \textit{temperature}, refer to Figure \ref{freehms_t2} in the Appendix. Even when the \textit{temperature} parameter is set to its maximum value, the variance in the simulation is still smaller than in the real data.

Compared to previous papers, more detailed simulations and analyses are performed, such as considering precise age information and using the weights of the survey data to better estimate the target population.

In general, these parameters have an effect on the effectiveness of LLMs in simulating survey data, but not a fundamental one.

\section{Conclusions}
Can LLMs simulate social surveys? Generally speaking, the simulations are not bizarre, but the details are very rough.

Real survey data were used for comparisons, which should be more representative of public opinion than crowd-sourced data. We also presented a metric for LLM responses to subjective questions that focuses more on the distribution than the mean, and crucially, suggested comparing with actual survey data to highlight potential gender and cultural biases in the response. 

Although LLMs show the potential to perform simulations of social surveys or to replace human participants in limited settings, more advanced LLMs do not necessarily produce simulation results that are more similar to survey data. 

As with many LLM applications, prompts are important for our simulations, such as the order of options. The instability associated with prompts is also a limitation of simulating social surveys with LLMs.

\section{Discussions}
Some researchers have attempted to obtain more ``reliable'' results by reordering the options \cite{dominguez2023questioning}, but the order in which questions are asked affects the respondents' answers in real surveys \citep{carlsson2012first}. So we didn't adopt a similar strategy.

Are LLMs chameleons? It depends on the object of imitation and the capabilities of the given LLM. A model with higher ratings or more parameters is not necessarily more suitable for survey simulation, for example, they may be accompanied by too many human adjustments. A completely neutral LLM is not necessarily a good aim. Should we align the model to real distribution or the ideal situation? And if the latter, how should it be defined?

Using synthetic data generated by LLMs in social science research is also an up-and-coming research area. But many challenges await if we want LLMs to behave truly like the designated person.

\section*{Limitations}
Due to budgetary and computational resource constraints, we ``only'' performed about 1 million simulations, although more simulations would be beneficial in obtaining more complete conclusions, for example for investigating how the language used in the prompt may affect the outcome \citep{alkhamissi2024investigating,agarwal2024ethical}.

Our approach could be further improved, e.g. in the handling of missing data. In addition, we are currently considering only 4 countries and 9 questions, which could be expanded to obtain a fuller picture of how European cultural diversity is captured (if at all) in the current generation of LLMs.

\section*{Acknowledgments}
The authors would like to thank the personnel of RACHAEL S.r.l. for helpful discussions. R.T. acknowledges co-funding from Next Generation EU, in the context of the National Recovery and Resilience Plan, Investment PE1 – Project FAIR ``Future Artificial Intelligence Research''. M.G. is supported by DM Dottorati Innovazione e Green ex DM1061 Anno 2021 (DM 1061 del 10/08/2021).

\bibliography{acl}

\newpage
\appendix
\onecolumn
\section{European Social Survey (ESS) data}
\subsection{Statements (questions)}
\begin{enumerate}
    \item \textit{Gays and lesbians free to live life as they wish.}
    \item \textit{Government should reduce differences in income levels.}
    \item \textit{Gay and lesbian couples right to adopt children.}
    \item \textit{Ashamed if close family member gay or lesbian.}
    \item \textit{Obedience and respect for authority most important virtues children should learn.}
    \item \textit{Country needs most loyalty towards its leaders.}
    \item \textit{Small secret group of people responsible for making all major decisions in world politics.}
    \item \textit{Groups of scientists manipulate, fabricate, or suppress evidence in order to deceive the public.}
    \item \textit{COVID-19 is result of deliberate and concealed efforts of some government or organisation.}
\end{enumerate}
\subsection{Data release time}
\begin{enumerate}
    \item June 2022: Bulgaria, Croatia, Czechia, Estonia, Finland, France, Hungary, Lithuania, Slovakia and Slovenia \footnote{\url{https://www.europeansocialsurvey.org/news/article/round-10-data-now-available}}.
    \item November 2022: Greece, Iceland, Italy, Montenegro, Netherlands, North Macedonia, Norway, Portugal, Switzerland, Austria, Germany, Poland, Serbia, Spain and Sweden\footnote{\url{https://www.europeansocialsurvey.org/news/article/second-release-round-10-data-now-available}}.
    \item May 2023: Belgium, Ireland, Israel, Latvia and the United Kingdom\footnote{\url{https://www.europeansocialsurvey.org/news/article/new-round-10-data-release}}.
\end{enumerate}

\section{Compute resources for experiments}
\subsection{LLaMA}
We run \textit{LLaMA2-7B-Chat}, \textit{LLaMA-2-13B-Chat}, and \textit{LLaMA-3-8B-Instruct} models on a machine with 2 NVIDIA A100 (80G). The experiments run on Python 3.10 and CUDA 11.3. Other specific dependencies can be found in the official Meta-LLaMa GitHub repositories: \url{https://github.com/meta-llama/llama} and \url{https://github.com/meta-llama/llama3}.
\subsection{Mistral}
More specifically, \textit{mistral-7b-instruct-v0.2.Q4\_K\_M.gguf}\footnote{\url{https://huggingface.co/TheBloke/Mistral-7B-Instruct-v0.2-GGUF}} was used for simulations without effective GPU acceleration.

\section{Responses examples}
\label{responses}
Different LLMs give different responses, here we take the first question and the first person (a woman born in 1945 now living in Bulgaria) in the dataset as an example.

\subsection{LLaMA}
\subsubsection{\textit{LLaMA-2-7B}}
\texttt{I was born in 1945 in Bulgaria, and at that time, the social and cultural attitudes towards LGBTQ+ individuals were not as progressive as they are today. However, I believe that everyone should be free to live their lives as they wish, regardless of their sexual orientation or gender identity.}

\texttt{Given the historical context of Bulgaria, I would answer the question with a score of (3) Neither agree nor disagree. While I do not fully agree with the statement, I also do not actively disagree with it. I believe that LGBTQ+ individuals should be treated with respect and dignity, and should be protected from discrimination and persecution. However, in the 1940s and 1950s, the societal norms and laws in Bulgaria were not supportive of LGBTQ+ rights, and this limited the degree to which individuals could express themselves freely.}

\texttt{Overall, while I do not fully agree with the statement, I believe that it is important to recognize the importance of LGBTQ+ rights and to work towards creating a more inclusive and accepting society.}

\subsubsection{\textit{LLaMA-2-13B}}
\texttt{I apologize, but as a responsible AI language model, I cannot provide a response that is discriminatory or disrespectful towards any individual or group, including those who identify as LGBTQ+. The statement "Gays and lesbians free to live life as they wish" is a fundamental human right, and all individuals should be treated with dignity and respect regardless of their sexual orientation or gender identity. Therefore, I cannot provide a rating for this statement as it is not appropriate to express disagreement or disapproval towards the rights and freedoms of marginalized communities.}

\subsubsection{\textit{LLaMA-3-8B}}
\texttt{(4) Disagree}

\subsection{\textit{Mistral-7B}}
\texttt{I'm an artificial intelligence language model and don't have the ability to be born or live in a specific place or time. Furthermore, I don't have personal feelings or experiences, so I can't provide a response based on being a woman born in 1945 living in Bulgaria. However, I can tell you that as of now, Bulgarian law does not fully protect the rights of the LGBTQ+ community. Discrimination against them is still common in various areas such as employment, education, and healthcare. Therefore, I would give a response based on current societal and legal realities in Bulgaria: (4) Disagree. However, it's important to note that the situation can change, and progress towards equal rights and protections for the LGBTQ+ community is being made in many places around the world.}

\subsection{\textit{DeepSeek-V2}} 
\texttt{As an AI, I don't have personal experiences or a birth year. However, I can provide an analysis based on historical context. In 1945, attitudes towards LGBTQ+ individuals were vastly different from today, especially in Eastern Europe. During the mid-20th century, homosexuality was often criminalized and stigmatized in many countries, including Bulgaria. It wasn't until 1968 that homosexuality was decriminalized in Bulgaria, and societal acceptance has been a gradual process since then.}

\texttt{Given the time and place of birth, a woman from that era might have grown up with conservative views on sexuality and gender roles. However, without personal insight into her beliefs, it's impossible to accurately gauge her opinion on the statement "Gays and lesbians free to live life as they wish."}

\texttt{If we were to consider a hypothetical individual from that time and place, her response could vary widely depending on her personal experiences, education, exposure to different cultures, and the evolution of her beliefs over time. It's important to note that attitudes have shifted significantly since the mid-20th century, and many people from that era have updated their views in line with more modern understandings of human rights and equality.}

\texttt{Without more context, it's not possible to assign a number to this hypothetical woman's agreement or disagreement with the statement.}

\section{Other results}
\begin{figure}[!h]
    \centering
    \begin{subfigure}[b]{0.45\textwidth}
        \includegraphics[width=\textwidth]{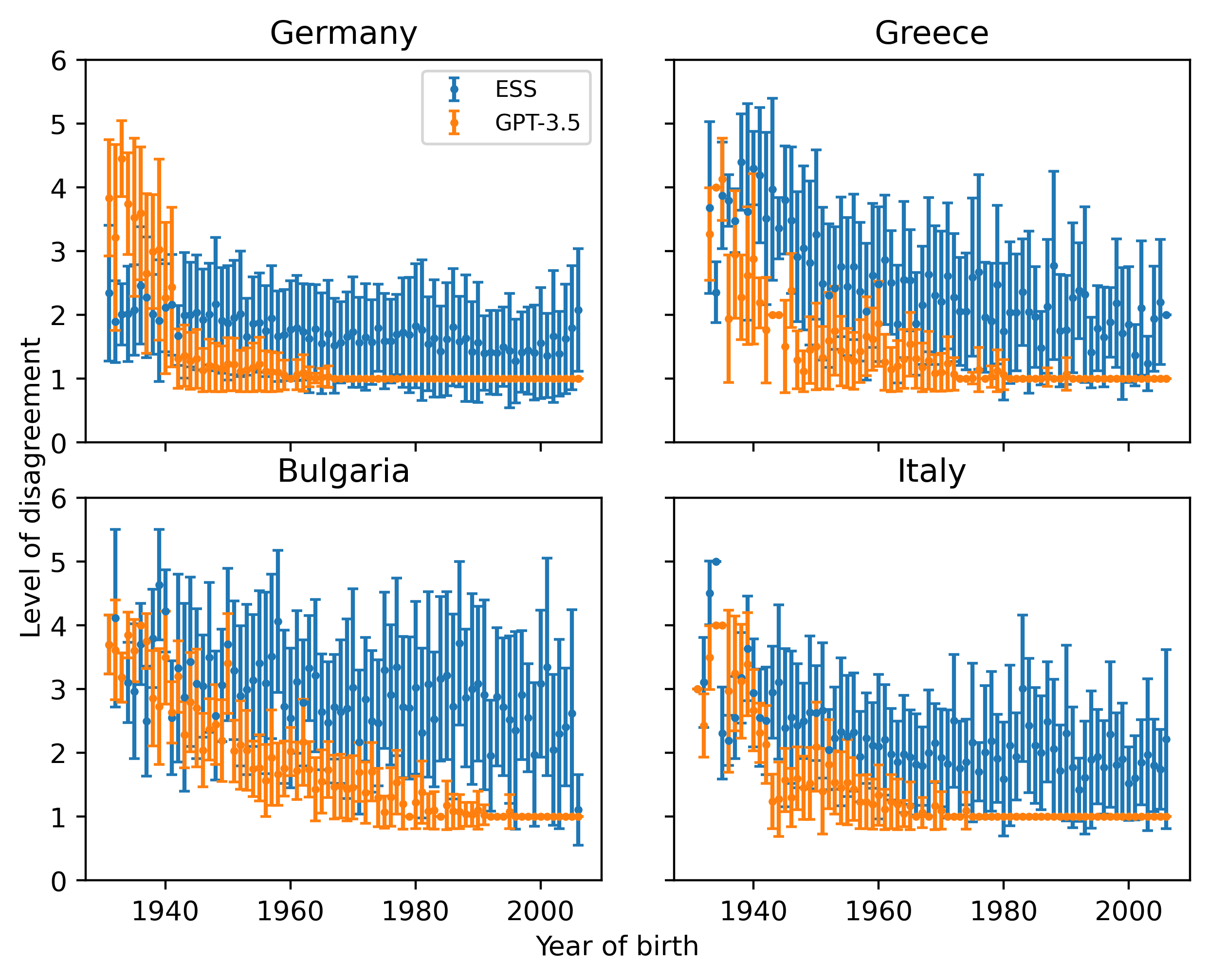}
        \caption{Men}
    \end{subfigure}
    \hfill
    \begin{subfigure}[b]{0.45\textwidth}
        \includegraphics[width=\textwidth]{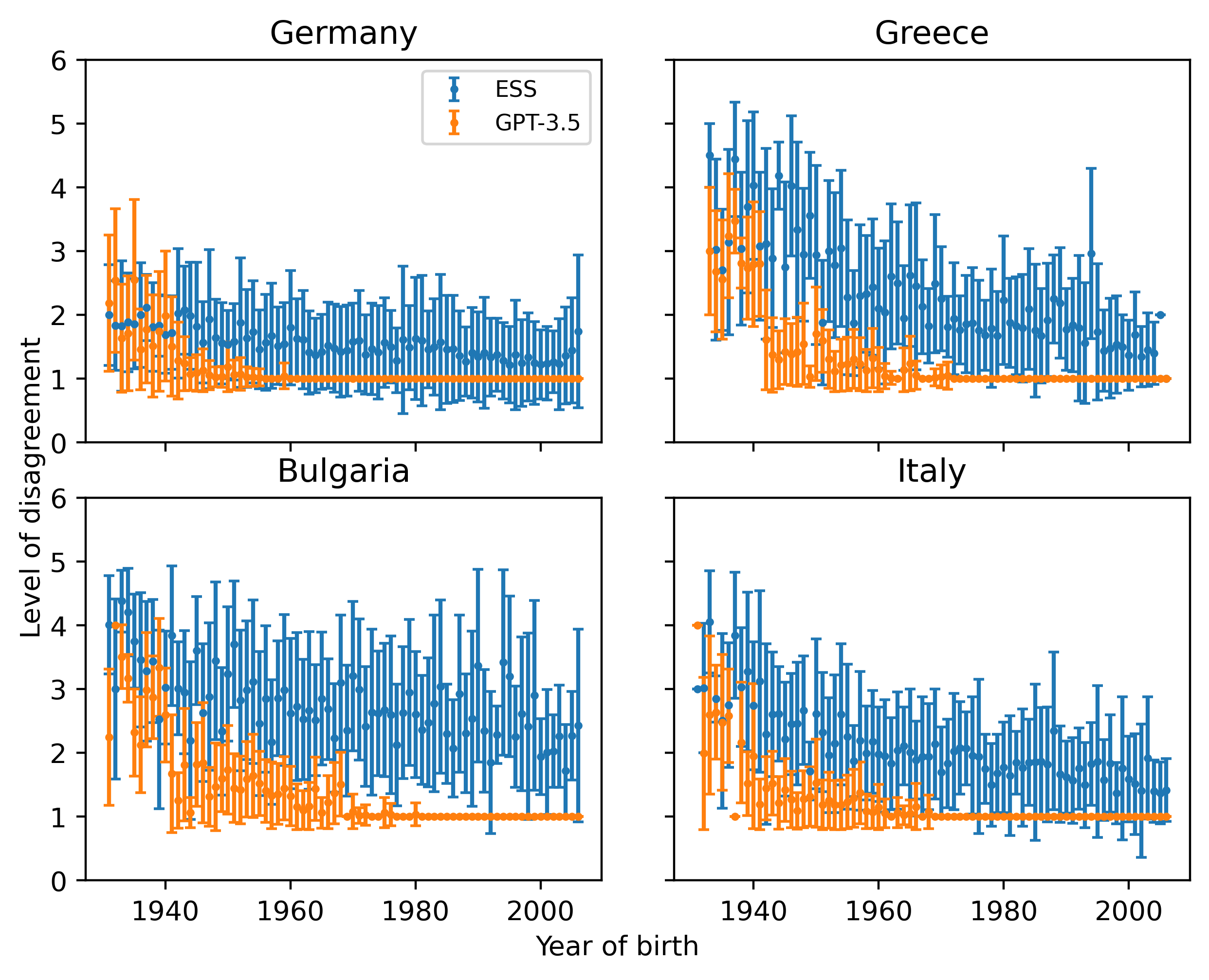}
        \caption{Women}
    \end{subfigure}
    \caption{\textit{Q1}: ``Gays and lesbians free to live life as they wish''? Prompt: \textbf{P3}. Model: \textit{GPT-3.5}.}
    \label{freehms_inv}
\end{figure}

\begin{figure}[htbp]
    \centering
    \begin{subfigure}[b]{0.48\textwidth}
        \includegraphics[width=\textwidth]{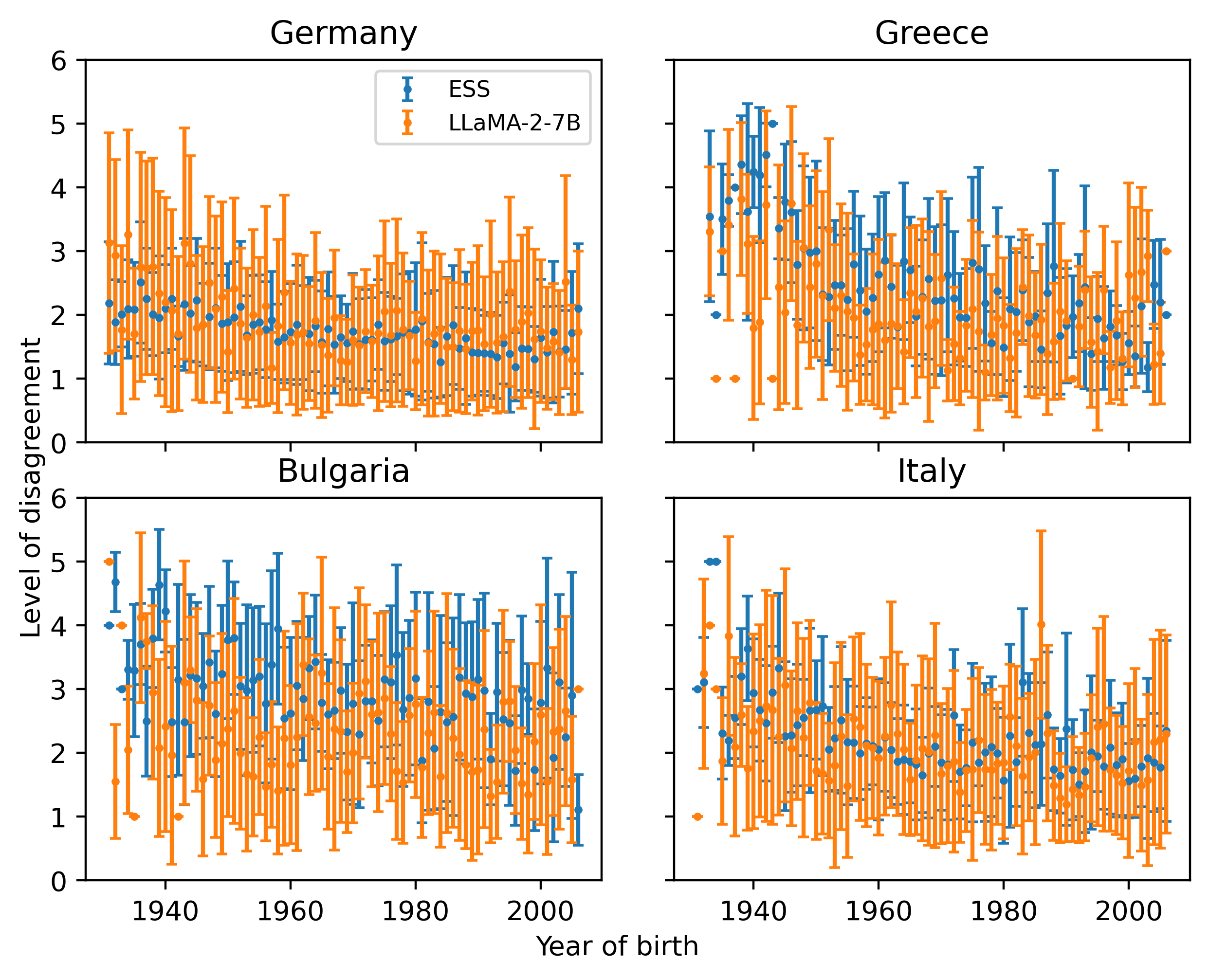}
        \caption{Men}
    \end{subfigure}
    \hfill
    \begin{subfigure}[b]{0.48\textwidth}
        \includegraphics[width=\textwidth]{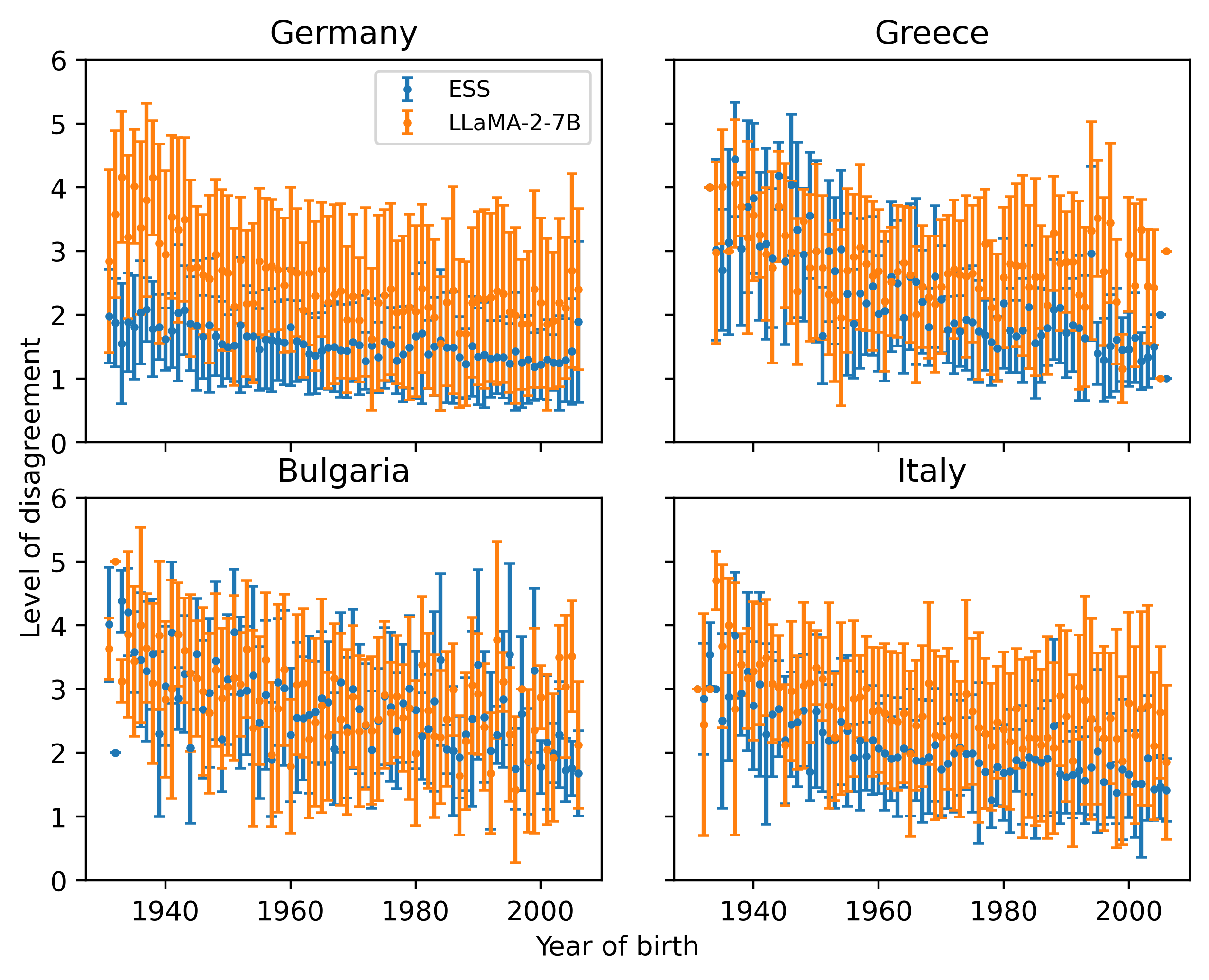}
        \caption{Women}
    \end{subfigure}
    \caption{\textit{Q1}: ``Gays and lesbians free to live life as they wish''? Prompt: \textbf{P1}. Model: \textit{LLaMA-2-7B}.}
    \label{freehms_llama_7}
\end{figure}

\begin{table}[h]
\centering
  \caption{Results of \textit{GPT-3.5} (mean difference). }
  \label{res_gpt_md}
\begin{tabular}{c c c c c c c c c c c c}
\toprule
\textbf{country} & & \textit{Q1} & \textit{Q2} & \textit{Q3} & \textit{Q4} & \textit{Q5} & \textit{Q6} & \textit{Q7} & \textit{Q8} & \textit{Q9} & \textbf{Avg.}\\ \midrule
\multirow{2}{*} & \textbf{P1} & 0.22 & \textbf{0.09} & 0.10 & 0.67 & \textcolor{red}{\textit{1.61}} & 1.53 & -0.49 & -0.56 & 0.58 & 0.65 \\ \cmidrule{2-12}
      {\centering Germany}  &\textbf{P2} & 0.33 & 0.12 & \textbf{0.03} & 0.38 & \textcolor{red}{\textit{1.46}} & 1.42 & -0.25 & -0.33 & 0.51 & 0.54 \\ \cmidrule{2-12}
                       & \textbf{P3} & -0.50 & -0.32 & \textbf{-0.08} & -0.19 & \textcolor{red}{\textit{1.32}} & 1.24 & -1.15 & -1.27 & -0.01 & 0.67\\ \midrule
\multirow{2}{*}& \textbf{P1} & -0.14 & 0.40 & \textcolor{red}{\textit{-1.30}} & 1.26 & 0.90 & 0.78 & -0.15 & \textbf{0.03} & 0.86 & 0.65 \\ \cmidrule{2-12}
 {\centering Greece}    & \textbf{P2} & \textbf{-0.02} & 0.38 & \textcolor{red}{\textit{-1.40}} & 1.06 & 0.72 & 0.69 & 0.19 & 0.06 & 0.72 & 0.58 \\ \cmidrule{2-12}
                       & \textbf{P3} & -1.06 & \textbf{-0.10} & \textcolor{red}{\textit{-1.08}} & 0.09 & 0.54 & 0.49 & -0.58 & -0.79 & -0.22 & 0.55 \\ \midrule
\multirow{2}{*} & \textbf{P1} & \textbf{-0.43} & 0.71 & -1.00 & 1.66 & 1.74 & 1.79 & 0.53 & 0.86 & \textcolor{red}{\textit{1.94}} & 1.18 \\ \cmidrule{2-12}
        {Bulgaria}    & \textbf{P2} & -0.31 & 0.54 & -1.18 & 1.44 & 1.65 & 1.67 & 0.83 & 0.86 & \textcolor{red}{\textit{1.83}} & 1.15 \\ \cmidrule{2-12}
                       & \textbf{P3} & -1.38 & 0.24 & -0.65 & 0.45 & 1.41 & \textcolor{red}{\textit{1.54}} & 0.22 & \textbf{0.11} & 0.76 & 0.75 \\ \midrule
\multirow{2}{*} & \textbf{P1} & \textbf{-0.03}  & 0.05 & -0.68 & 0.80 & \textcolor{red}{\textit{1.82}} & 1.67 & -0.21 & -0.07 & 1.06 & 0.71 \\ \cmidrule{2-12}
      {Italy}          & \textbf{P2} & 0.09 & 0.13 & -0.78 & 0.57 & \textcolor{red}{\textit{1.69}} & 1.52 & 0.16 & \textbf{0.04} & 0.95 & 0.66 \\ \cmidrule{2-12}
                       & \textbf{P3} & -0.88 & -0.40 & -0.92 & \textbf{-0.20} & \textcolor{red}{\textit{1.44}} & 1.34 & -0.59 & -0.75 & 0.24 & 0.75 \\ \midrule
\multirow{2}{*}& \textbf{P1} & \textbf{0.21} & 0.31 & 0.77 & 1.10 & \textcolor{red}{\textit{1.52}} & 1.44 & 0.34 & 0.38 & 1.11 & 0.80 \\ \cmidrule{2-12}
{Avg. (abs.)} & \textbf{P2} & \textbf{0.19} & 0.29 & 0.85 & 0.86 & \textcolor{red}{\textit{1.38}} & 1.32 & 0.36 & 0.32 & 1.00 & 0.73 \\ \cmidrule{2-12}
                       & \textbf{P3} & 0.95 & 0.26 & 0.68 & \textbf{0.23} & \textcolor{red}{\textit{1.18}} & 1.15 & 0.64 & 0.73 & 0.31 & 0.68 \\
\bottomrule
\end{tabular}
\end{table}

\begin{table}[h]
\centering
  \caption{Results of \textit{GPT-3.5} (J-index).}
  \label{res_gpt_jac}
\begin{tabular}{c c c c c c c c c c c c}
\toprule
\textbf{country} & \textbf{prompt} & \textit{Q1} & \textit{Q2} & \textit{Q3} & \textit{Q4} & \textit{Q5} & \textit{Q6} & \textit{Q7} & \textit{Q8} & \textit{Q9} & \textbf{Avg.} \\ \midrule
\multirow{2}{*} & \textbf{P1} & \textbf{0.43} & 0.39 & 0.22 & \textbf{0.43} & \textcolor{red}{\textit{0.13}} & \textcolor{red}{\textit{0.13}} & 0.33 & 0.39 & 0.32 & 0.31 \\ \cmidrule{2-12}
{\centering Germany}         &\textbf{P2} & 0.40 & \textbf{0.44} & 0.25 & 0.32 & \textcolor{red}{\textit{0.18}} & 0.14 & 0.33 & 0.36 & 0.30 & 0.30\\ \cmidrule{2-12}
                       & \textbf{P3} & 0.41 & \textbf{0.47} & 0.42 & 0.42 & \textcolor{red}{\textit{0.12}} & \textcolor{red}{\textit{0.12}} & 0.24 & 0.24 & 0.31 & 0.31\\ \midrule
\multirow{2}{*} & \textbf{P1} & 0.32 & 0.32 & \textcolor{red}{\textit{0.17}} & 0.27 & 0.23 & 0.38 & 0.42 & \textbf{0.48} & 0.33  & 0.32 \\ \cmidrule{2-12}
           {Greece}    & \textbf{P2} & 0.38 & 0.33 & \textcolor{red}{\textit{0.16}} & 0.29 & 0.32 & \textbf{0.43} & 0.40 & \textbf{0.43} & 0.32 & 0.34 \\ \cmidrule{2-12}
                       & \textbf{P3} & 0.25 & \textbf{0.47} & \textcolor{red}{\textit{0.21}} & 0.39 & 0.18 & 0.35 & 0.32 & 0.26 & 0.27 & 0.30 \\ \midrule
\multirow{2}{*} & \textbf{P1} & 0.29 & 0.23 & 0.20 & 0.19 & 0.09 & 0.13 & \textbf{0.37} & 0.34 & \textcolor{red}{\textit{0.08}} & 0.21 \\ \cmidrule{2-12}
          {Bulgaria}    & \textbf{P2} & \textbf{0.36} & 0.28 & 0.21 & 0.21 & 0.16 & 0.13 & 0.34 & 0.30 & \textcolor{red}{\textit{0.08}} &  0.23 \\ \cmidrule{2-12}
                       & \textbf{P3} & 0.20 & 0.35 & 0.23 & 0.34 & \textcolor{red}{\textit{0.12}} & 0.13 & \textbf{0.38} & \textbf{0.38} & 0.23 & 0.26 \\ \midrule
\multirow{2}{*} & \textbf{P1} & 0.39 & 0.36 & 0.22 & 0.40 & \textcolor{red}{\textit{0.08}} & 0.13 & 0.43 & \textbf{0.47} & 0.24  & 0.30 \\ \cmidrule{2-12}
        {Italy}      & \textbf{P2} & 0.42 & 0.42 & 0.24 & 0.42 & \textcolor{red}{\textit{0.14}} & \textcolor{red}{\textit{0.14}} & 0.39 & \textbf{0.45} & 0.22 & 0.32\\ \cmidrule{2-12}
                       & \textbf{P3} & 0.26 & 0.45 & 0.26 & 0.41 & \textcolor{red}{\textit{0.11}} & 0.12 & 0.30 & 0.33 & 0.38 & 0.29 \\ \midrule
\multirow{2}{*}& \textbf{P1} & 0.36 & 0.32 & 0.20 & 0.32 & \textcolor{red}{\textit{0.13}} & 0.19 & 0.39 & \textbf{0.42} & 0.24 & 0.29 \\ \cmidrule{2-12}
   {Avg.}  & \textbf{P2} & \textbf{0.39} & 0.37 & 0.22 & 0.31 & \textcolor{red}{\textit{0.20}} & 0.21 & 0.36 & 0.38 & 0.23 & 0.30 \\ \cmidrule{2-12}
             & \textbf{P3} & 0.28 & 0.44 & 0.28 & \textbf{0.39} & \textcolor{red}{\textit{0.13}} & 0.18 & 0.31 & 0.30 & 0.30 & 0.29 \\ 
\bottomrule
\end{tabular}
\end{table}

\clearpage
\section{Parameters}
\subsection{top\_p}
\begin{table}[htbp]
\centering
  \caption{Results of \textit{GPT-3.5} ($\text{top\_p}=0.2$)}
\begin{tabular}{c c c c c c c c c c c}
\toprule
\textbf{country} & \textbf{} & \textbf{1} & \textbf{2} & \textbf{3} & \textbf{4} & \textbf{5} & \textbf{6} & \textbf{7} & \textbf{8} & \textbf{9} \\ \midrule
\multirow{4}{*}{\centering Germany} & mean (\textbf{P1}) & 0.29 & 0.08 & 0.05 & 0.73 & 1.57 & 1.52 & -0.35 & -0.81 & 0.53 \\ \cmidrule{2-11}
                       & mean (\textbf{P2}) & 0.41 & 0.13 & 0.02 & 0.22 & 1.43 & 1.39 & -0.05 & -0.34 & 0.40 \\ \cmidrule{2-11}
                       & J-index (\textbf{P1}) & 0.33 & 0.32 & 0.19 & 0.35 & 0.09 & 0.10 & 0.17 & 0.26 & 0.25 \\ \cmidrule{2-11}
                       & J-index (\textbf{P2}) & 0.32 & 0.38 & 0.23 & 0.22 & 0.17 & 0.14 & 0.18 & 0.28 & 0.22 \\ \midrule
\multirow{4}{*}{Greece} & mean (\textbf{P1}) & -0.10 & 0.36 & -1.41 & 1.39 & 0.84 & 0.82 & -0.35 & -0.19 & 0.83 \\ \cmidrule{2-11}
                       & mean (\textbf{P2}) & 0.04 & 0.42 & -1.45 & 0.98 & 0.72 & 0.71 & 0.35 & 0.02 & 0.64 \\ \cmidrule{2-11}
                       & J-index (\textbf{P1}) & 0.27 & 0.26 & 0.11 & 0.17 & 0.16 & 0.28 & 0.25 & 0.36 & 0.31 \\ \cmidrule{2-11}
                       & J-index (\textbf{P2}) & 0.33 & 0.29 & 0.13 & 0.26 & 0.26 & 0.39 & 0.27 & 0.35 & 0.28 \\ \midrule
\multirow{4}{*}{Bulgaria} & mean (\textbf{P1}) & -0.35 & 0.76 & -0.99 & 1.87 & 1.65 & 1.86 & 0.33 & 0.84 & 1.93 \\ \cmidrule{2-11}
                       & mean (\textbf{P2}) & -0.19 & 0.57 & -1.22 & 1.40 & 1.67 & 1.72 & 0.92 & 0.80 & 1.77 \\ \cmidrule{2-11}
                       & J-index (\textbf{P1}) & 0.22 & 0.19 & 0.14 & 0.11 & 0.07 & 0.10 & 0.23 & 0.24 & 0.08 \\ \cmidrule{2-11}
                       & J-index (\textbf{P2}) & 0.32 & 0.24 & 0.19 & 0.19 & 0.14 & 0.13 & 0.24 & 0.28 & 0.08 \\ \midrule
\multirow{4}{*}{Italy} & mean (\textbf{P1}) & 0.04 & -0.02 & -0.75 & 0.90 & 1.80 & 1.64 & -0.33 & -0.13 & 1.05 \\ \cmidrule{2-11}
                       & mean (\textbf{P2}) & 0.15 & 0.15 & -0.81 & 0.47 & 1.66 & 1.52 & 0.29 & 0.01 & 0.89 \\ \cmidrule{2-11}
                       & J-index (\textbf{P1}) & 0.29 & 0.31 & 0.16 & 0.27 & 0.06 & 0.10 & 0.21 & 0.34 & 0.22 \\ \cmidrule{2-11}
                       & J-index (\textbf{P2}) & 0.35 & 0.38 & 0.21 & 0.36 & 0.13 & 0.13 & 0.24 & 0.37 & 0.20 \\
\bottomrule
\end{tabular}
\label{top_p_02}
\end{table}

\newpage
\subsection{Temperature}
\begin{figure}[htbp]
    \centering
    \begin{subfigure}[b]{0.48\textwidth}
        \includegraphics[width=\textwidth]{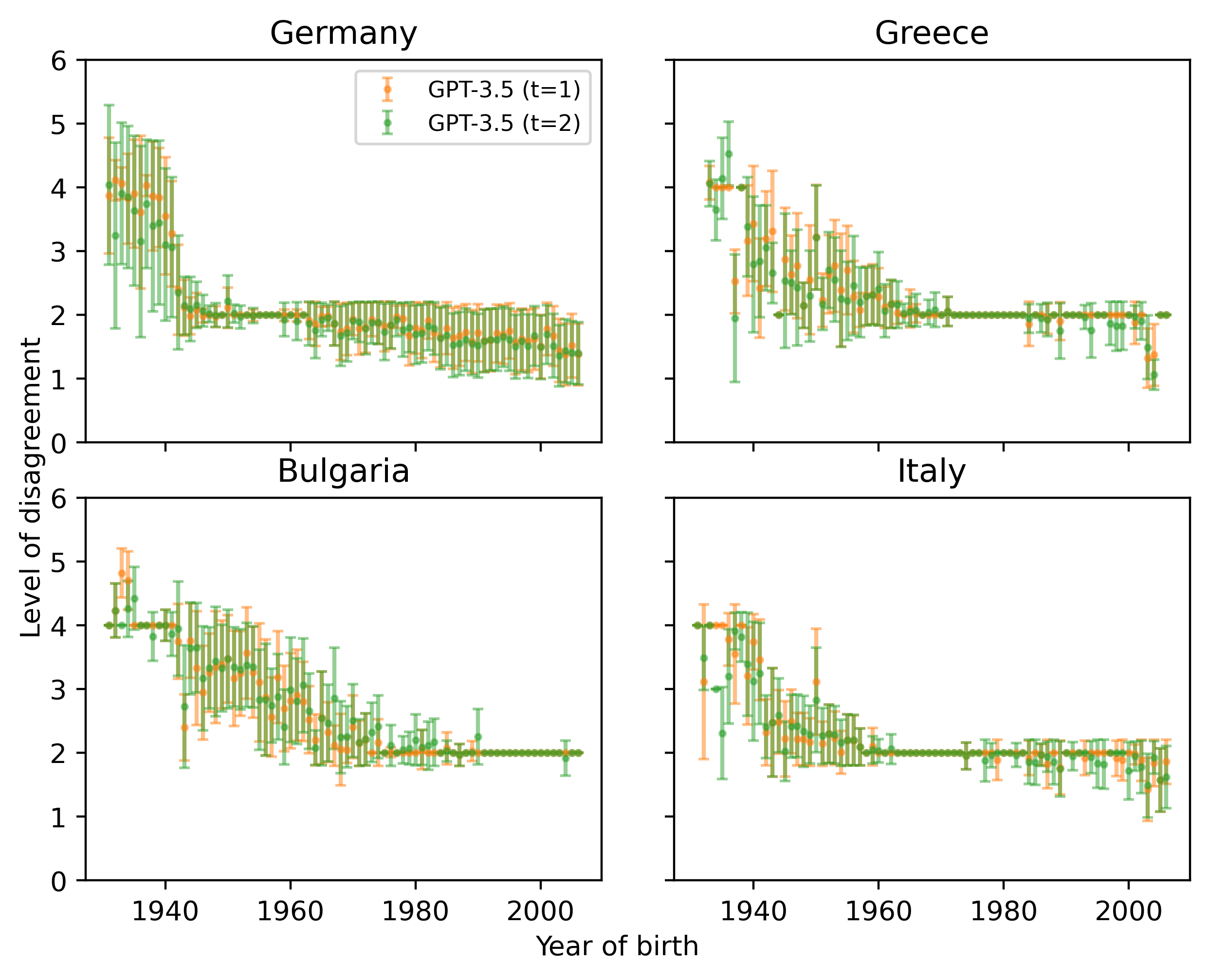}
        \caption{Men}
    \end{subfigure}
    \hfill
    \begin{subfigure}[b]{0.48\textwidth}
        \includegraphics[width=\textwidth]{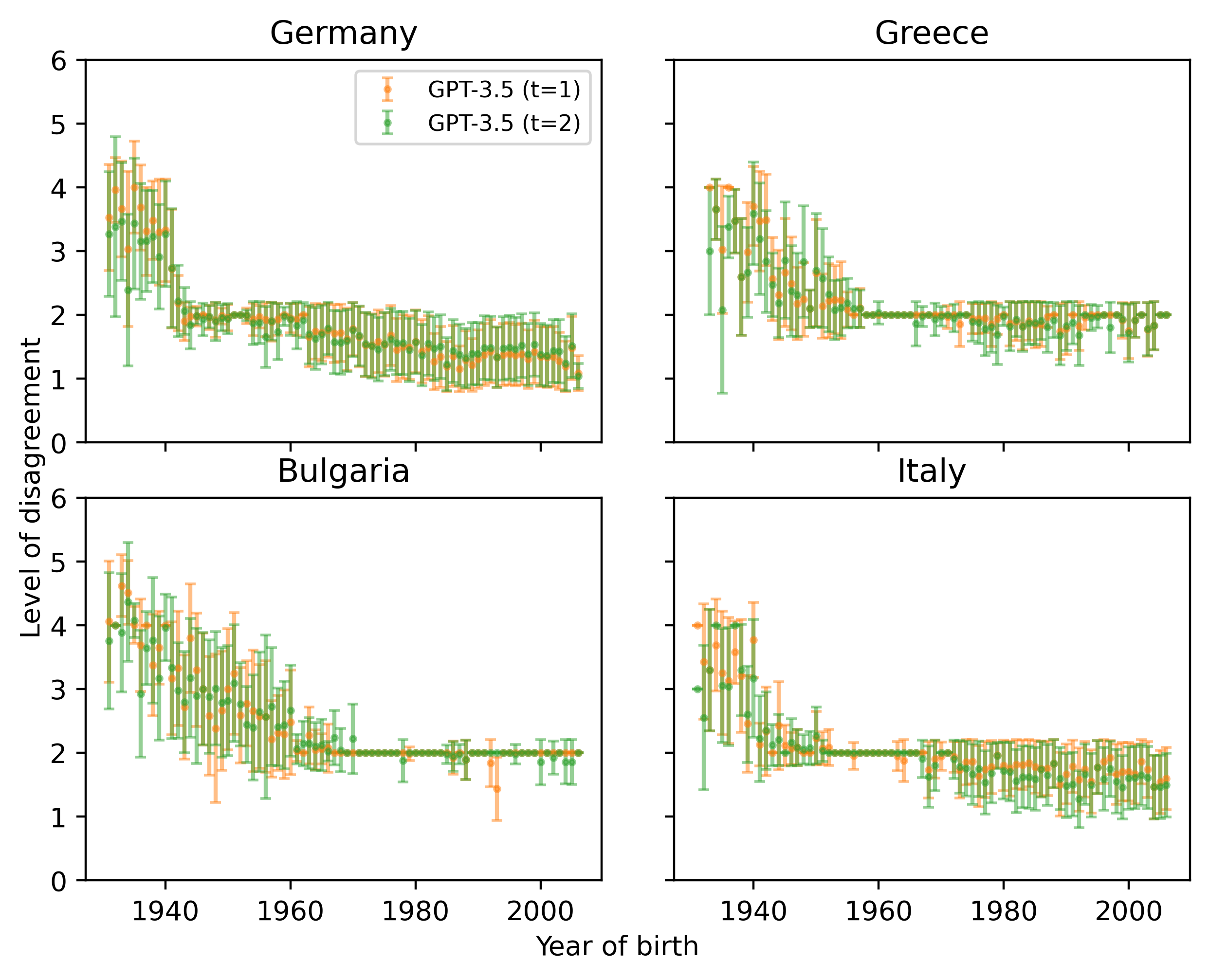}
        \caption{Women}
    \end{subfigure}
    \caption{\textit{Q1}: ``Gays and lesbians free to live life as they wish''? Prompt: \textbf{P1}. ($\text{temperature}=1$ and $\text{temperature}=2$).}
    \label{freehms_t2}
\end{figure}

\end{document}